\documentclass{article}
\usepackage[accepted]{icml2005}
\usepackage{epsf,psfig,wrapfig}

\newcommand{\capfont}[0]
  {\protect \footnotesize}

\begin{document}

\twocolumn [\icmltitle{Semi-Supervised Learning -- A Statistical Physics Approach}
\icmlauthor{Gad Getz$^{*\dag}$}{gaddy.getz@weizmann.ac.il} \icmlauthor{Noam
Shental$^{*}$}{noam.shental@weizmann.ac.il} \icmlauthor{Eytan
Domany}{eytan.domany@weizmann.ac.il} \icmladdress{Dept. of Physics
of Complex Systems, Weizmann Institute of Science, Rehovot 76100
Israel}
]


\begin{abstract}
We present a novel approach to semi-supervised learning  which is based on
 statistical physics.
Most of the former work in the field of semi-supervised learning
classifies the points by minimizing a certain energy function,
which corresponds to a minimal k-way cut solution.
In contrast to these methods, we estimate the {\em distribution} of classifications,
instead of the sole minimal k-way cut, which yields more accurate and robust
results.
Our approach may be applied to all energy functions used for semi-supervised learning.
The method is based on sampling using a Multicanonical Markov chain Monte-Carlo algorithm,
and has a straightforward probabilistic interpretation,
which allows for soft assignments of points to classes,
and also to cope with yet unseen class types.
The  suggested approach is  demonstrated on a toy data set and on two real-life
data sets of  gene expression.
\end{abstract}

\section{Introduction}
Situations in which  many unlabelled points are available  and  only few labelled points
are provided call for semi-supervised learning methods.
The goal of semi-supervised learning is to classify the unlabelled points,
on the basis of their distribution and the provided labelled points.
Such problems occur in many fields, in which obtaining data is cheap
but labelling is expensive.
In such scenarios supervised methods are impractical,
but the presence of the few labelled points can
significantly improve the performance of unsupervised methods.

The basic assumption of unsupervised learning, i.e. clustering,
is that points which belong to the same cluster actually originate
from the same class. Clustering methods which are based on
estimating the density of data points define a cluster as a `mode'
in the distribution, i.e. a relatively dense region
surrounded by  relatively lower density.
Hence each mode is assumed to originate from a single class,
although a certain class may be dispersed over several modes.

In case the modes are well separated  they can be easily identified by
unsupervised techniques, and there is no need for semi-supervised methods.
However, consider the case of two close  modes which belong to two different classes,
but the  density of points between them  is not significantly lower than the
density within each mode. In this case density based unsupervised methods may encounter difficulties
in distinguishing between  the modes (classes),
while semi-supervised methods can be of help.
Even if a few points are labelled in each class, semi-supervised
algorithms, which cannot cluster together points of different labels,
are {\em forced} to place a border between the modes.
Most probably the border will pass in between the modes,
where  the density of points is lower. Hence, the forced border
`amplifies' the otherwise less noticed differences between the modes.

\begin{figure*}
\begin{tabular}{cccccc}
\psfig{file=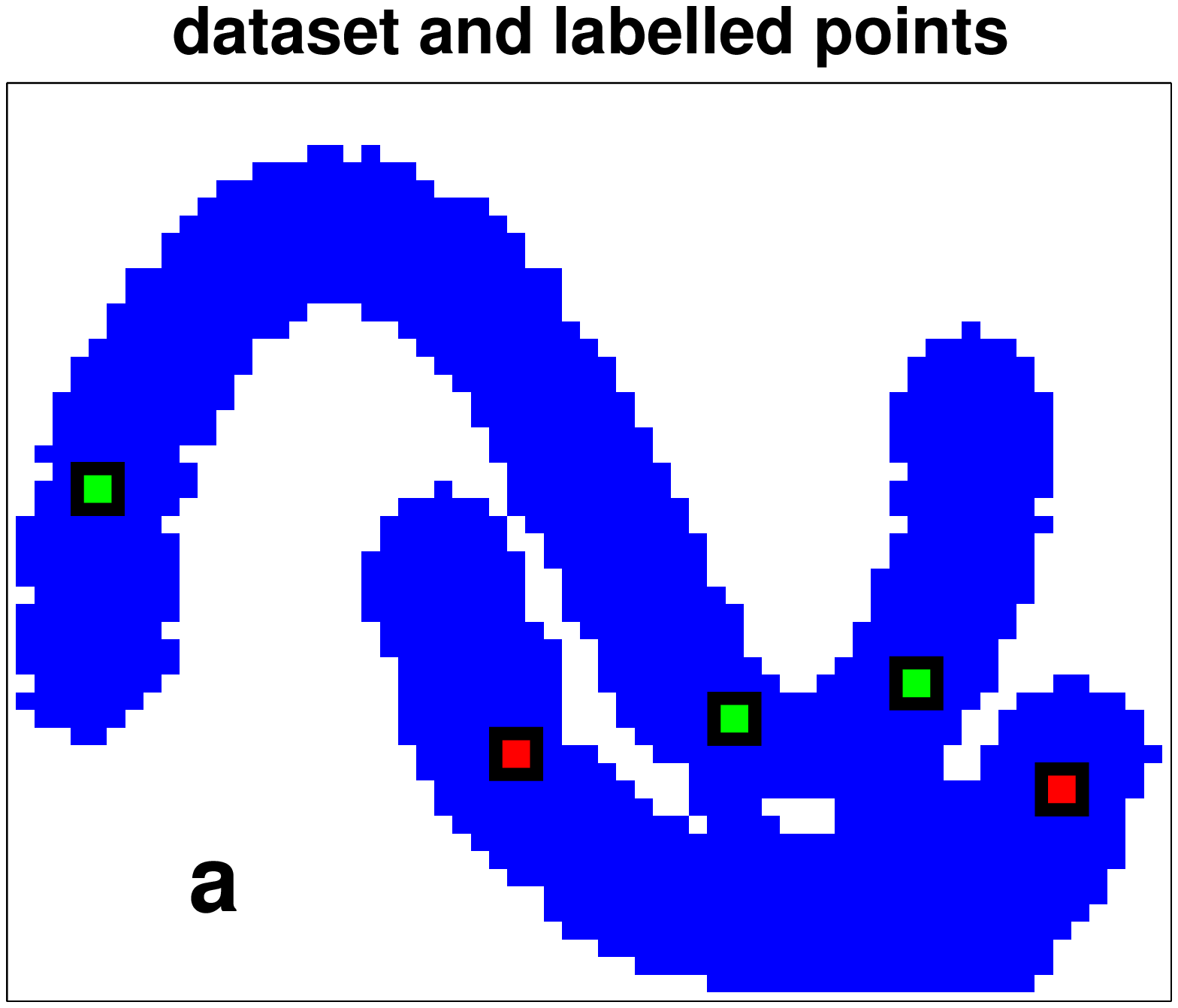,width=2.5cm} &
\hspace{-0.5cm}
\psfig{file=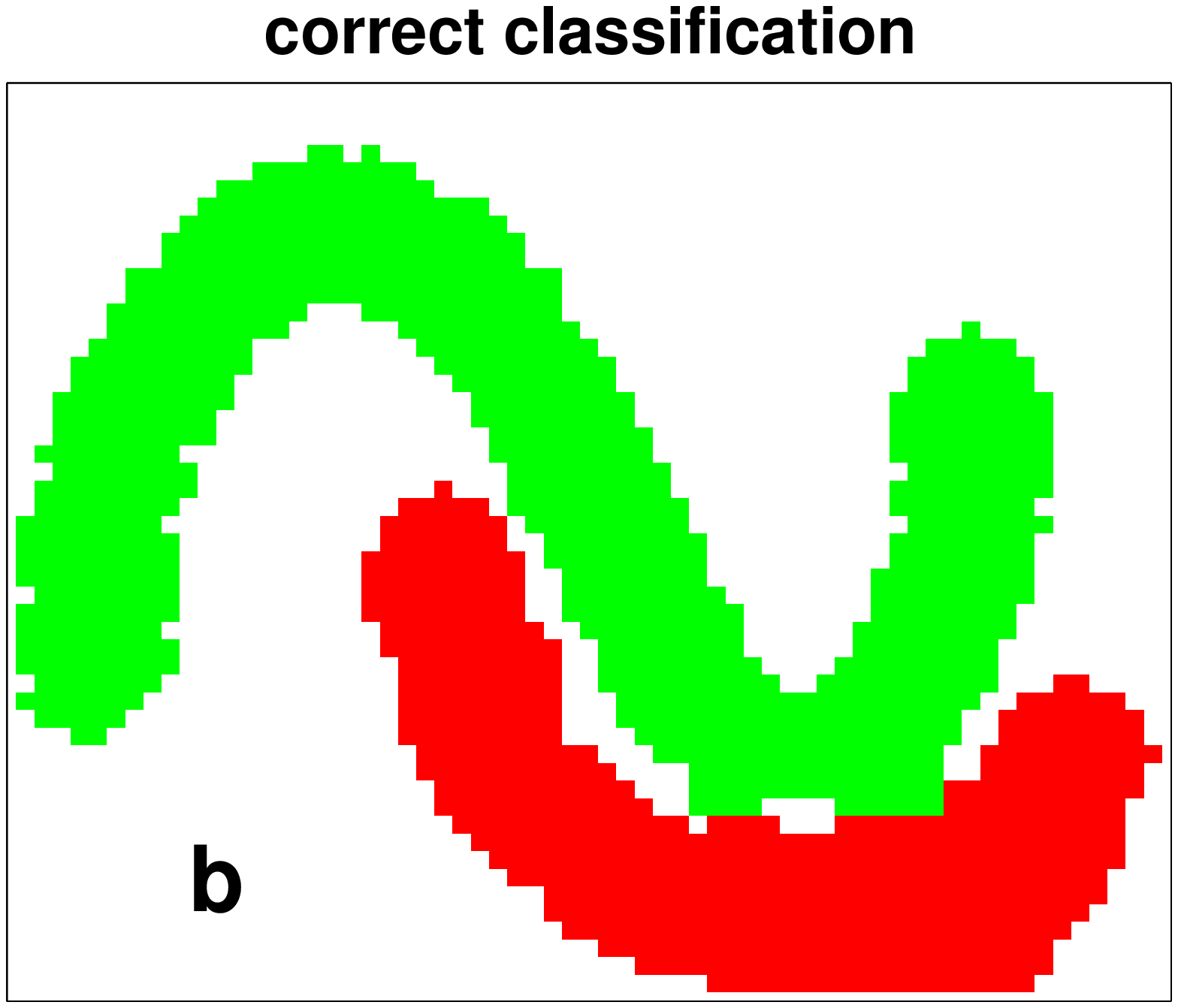,width=2.5cm} &
\hspace{-0.5cm}
\psfig{file=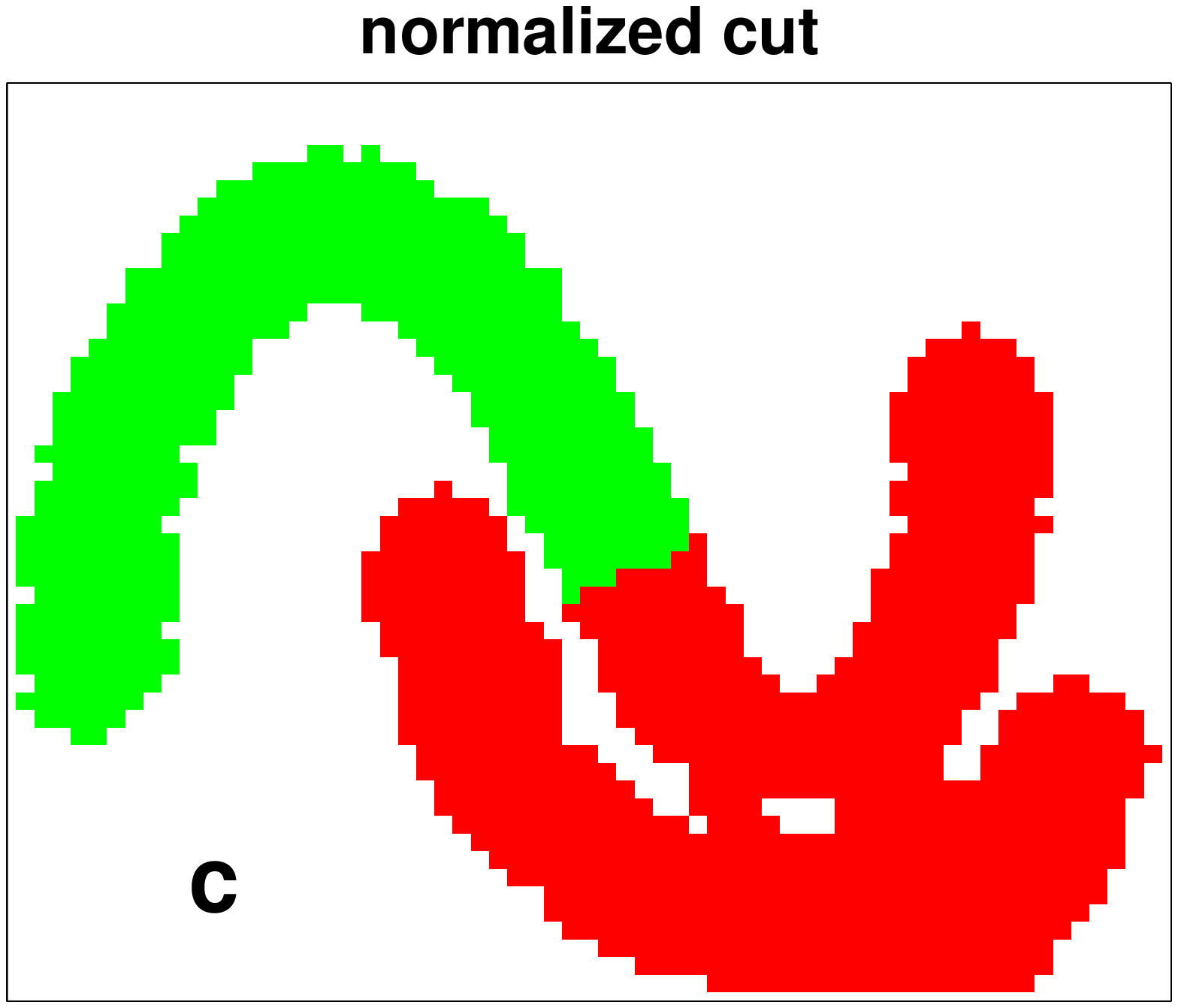,width=2.5cm} &
\hspace{-0.5cm}
\psfig{file=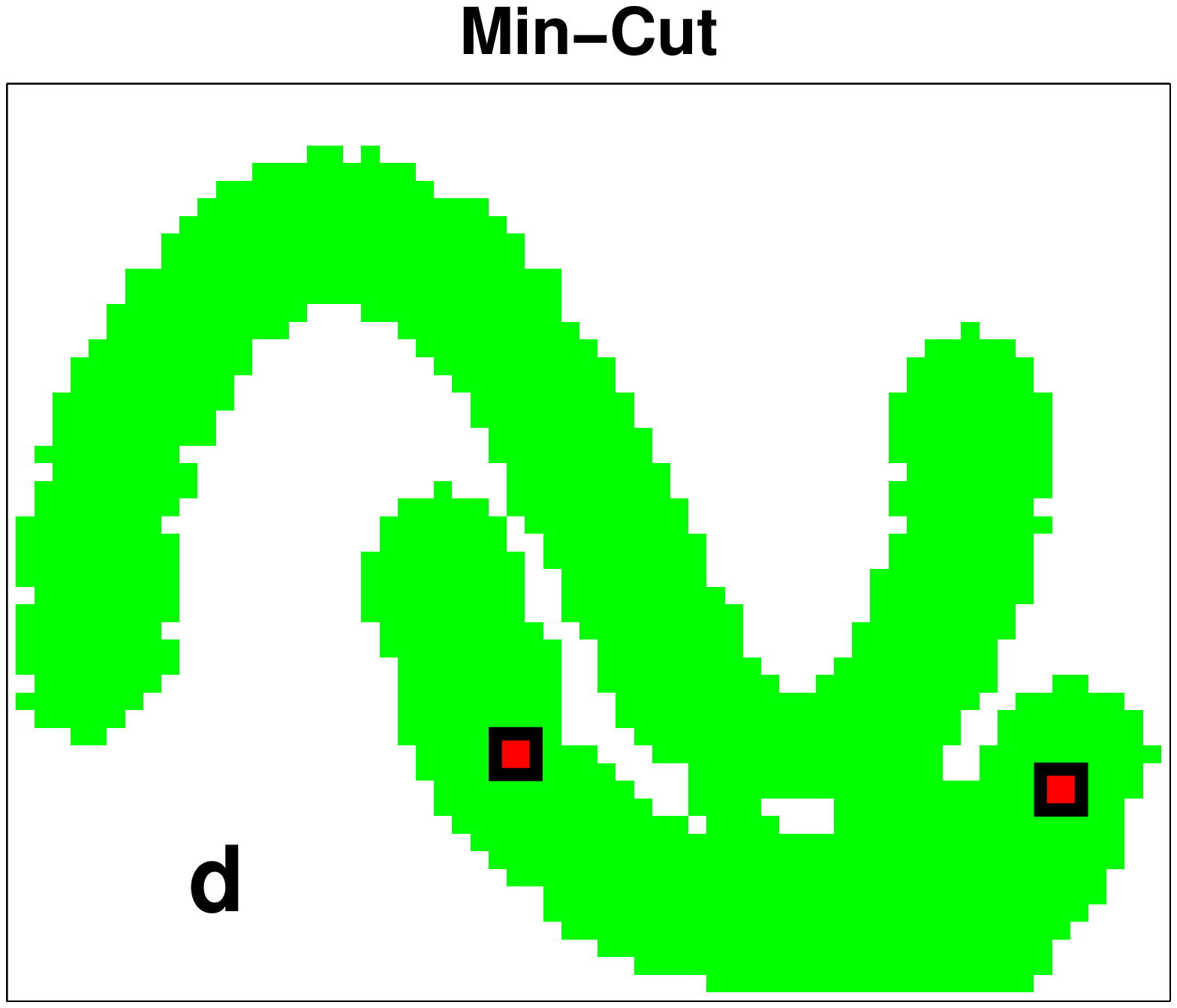,width=2.5cm} &
\hspace{-0.5cm}
\psfig{file=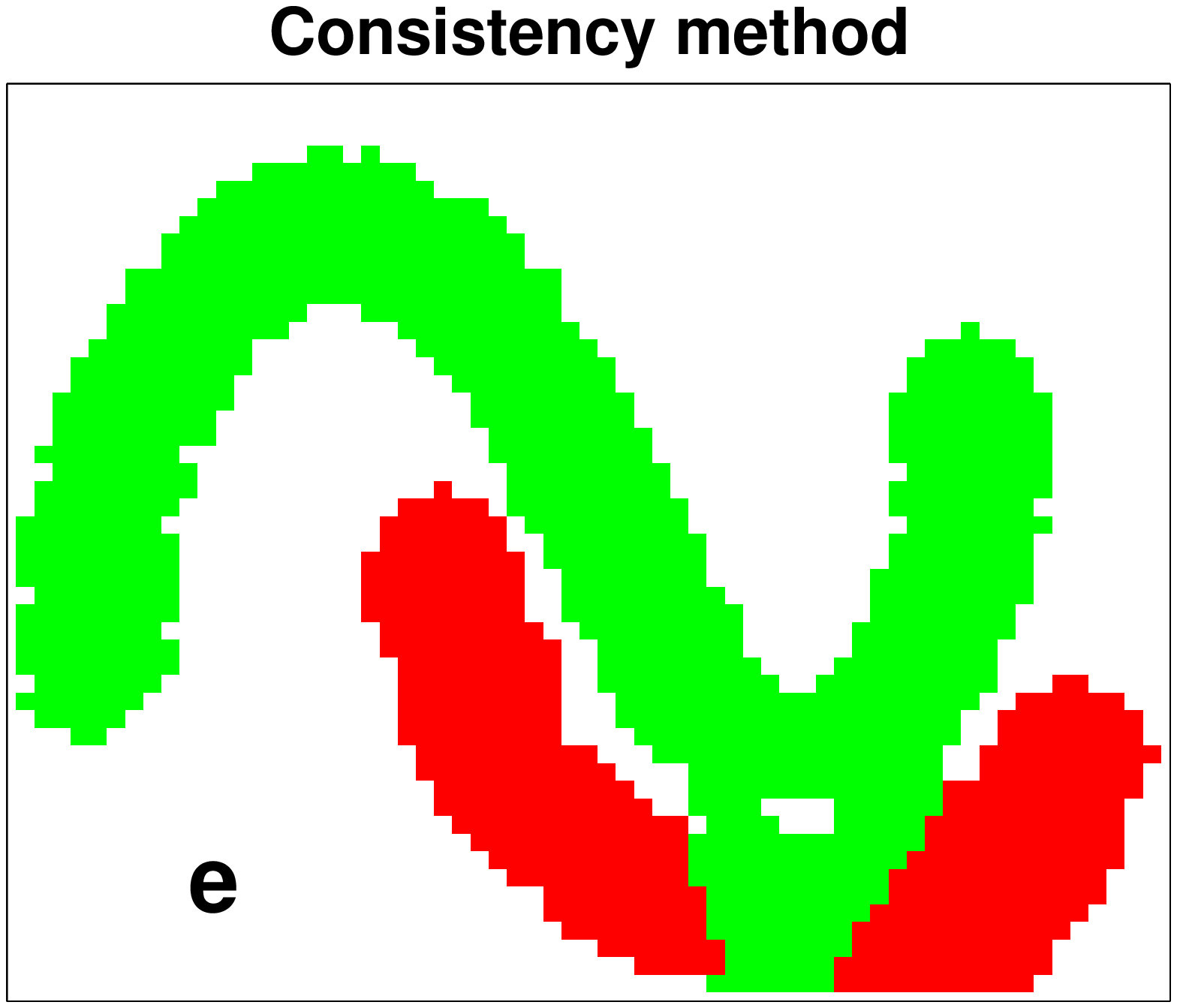,width=2.5cm} &
\hspace{-0.5cm}
\psfig{file=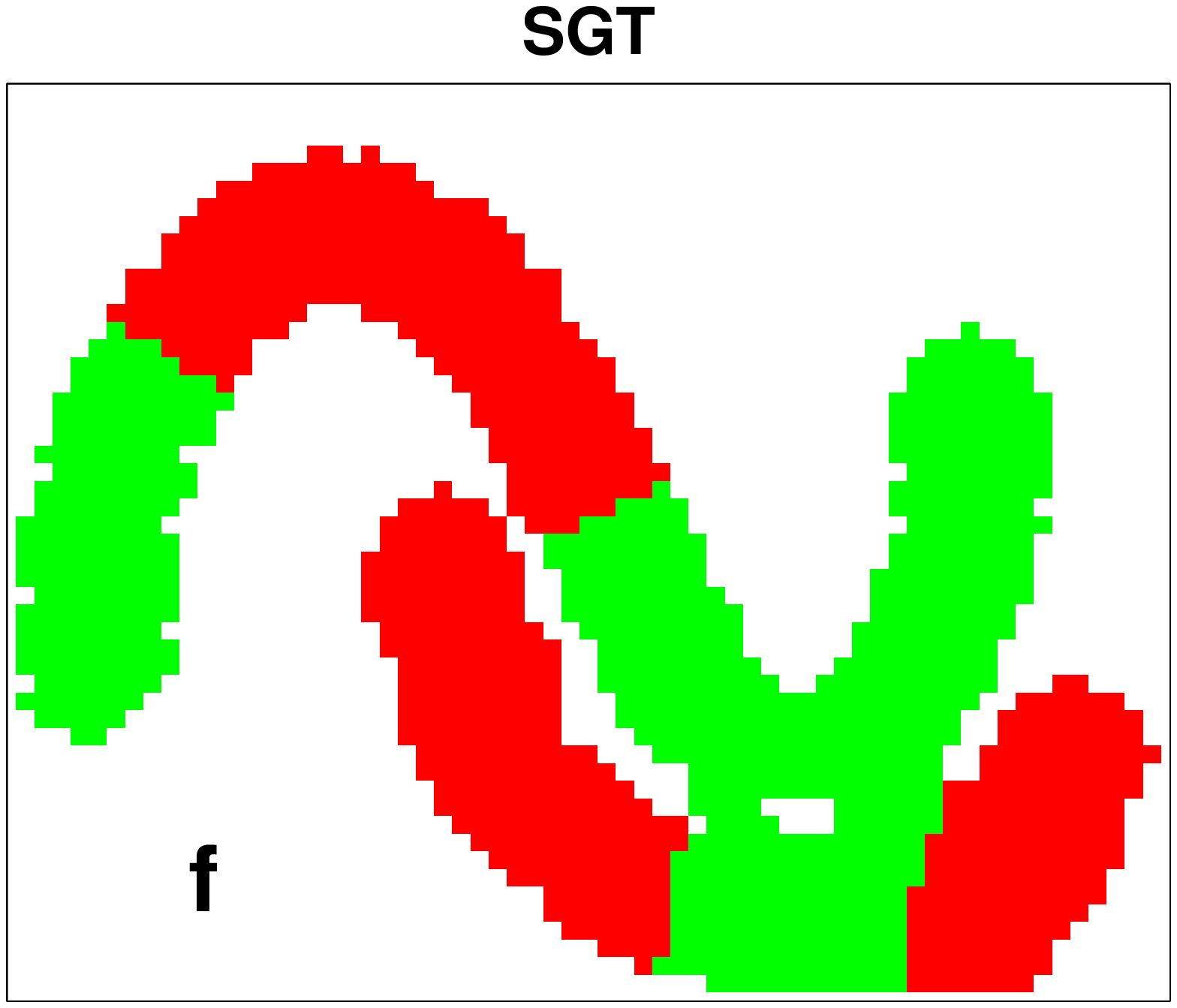,width=2.5cm}
\end{tabular}
\vspace{-0.25cm}
\caption{\capfont
a.  The unlabelled data in blue; labelled points are marked in green and in red.
Each of the $1360$ pixels correspond to a data point; the labeled pixels were enlarged for clarity.
b. The correct classification.
c. Clustering results of the unsupervised normalized cut algorithm~\cite{ShiMalik}.
d. The min-cut  solution.
e. The results of the semi-supervised consistency method  of~\cite{scholkopf}.
f. The outcome of the spectral graph transducer
  algorithm~\cite{sgt}, which is a semi-supervised extension of
  the normalized cut algorithm.
}
\label{fig:toy}
\vspace{-0.5cm}
\end{figure*}

For example, consider the image in Fig.~\ref{fig:toy}a.
Each pixel corresponds to a data point
and the similarity score between adjacent pixels is of value unity.
The green and red pixels are labelled while the rest of the blue pixels are unlabelled.
The desired classification into red and green classes appears in Fig.~\ref{fig:toy}b.
It is unlikely that any unsupervised method would partition the data correctly
(see e.g. Fig.~\ref{fig:toy}c) since the two classes   form  one uniform cluster.
However, using a few labelled points semi-supervised methods which must place a border between
the red and green classes may become useful.

In recent years various types of semi-supervised learning algorithms have been proposed,
however almost all of these methods share a common basic approach.
They define a certain cost function, i.e. {\em energy}, over the
possible classifications, try to minimize this energy, and output the
minimal energy classification as their solution.
Different methods vary by the specific energy function and by their minimization procedures;
for example the work on graph cuts~\cite{blum,boykov99},
minimizes the cost of a cut in the graph,
while others  choose to  minimize the normalized cut cost~\cite{sgt,yu},
or a quadratic cost~\cite{zhu03,scholkopf}.

As stated recently by~\cite{blumSemiSup}, searching for a minimal energy
has a  basic disadvantage, common to all former methods: it ignores the robustness of the found solution.
Blum et al. mention the case of  several minima with equal energy,
where  one arbitrarily chooses one solution, instead of considering them all.
Put differently, imagine the energy landscape in the space of  solutions;
it may contain many equal energy minima as considered Blum et al., but also other phenomena
may harm the robustness of the global minimum as an optimal  solution.
First,  it may happen that the difference in energy between the global minimum, and
close by solutions is minuscule, thus picking the minimum as the sole solution
may be incorrect or arbitrary.
Secondly, in many  cases there are too few data points (both labelled and unlabelled)
which may cause the empirical density to locally deviate from the true density.
Such fluctuations in the density may drive the
minimal energy solution far from the correct one.
For example, due to fluctuations a low density ``crack" may be formed inside
a high density region, which may erroneously split  a single  cluster in two.
Another type of fluctuation may generate  a ``filament" of high density
points in  a low density region, which may unite two clusters of different classes.
In both cases, the minimal energy solution is erroneously `guided' by the fluctuations,
and fails to find the correct classification.
An example of the latter case appears in Fig.~\ref{fig:toy}a;   the classifications
provided by three semi-supervised methods appear in Fig.~\ref{fig:toy}d--f,
fail to recover the desired classification, due to a `filament' which connects the  classes.

Searching for  the minimal energy solution is equivalent to seeking
the most probable joint classification (MAP).
A possible remedy to the difficulties in this approach may then be
to consider the probability distribution of all possible classifications.
Blum et al. provided a first step in this direction using a randomized min-cut  algorithm.
In this work we provide a different solution  based
on   statistical physics.

Basically each solution in our method  is weighed
by its energy $E(classification)$, also known as the Boltzmann weight,
and its probability is given by:
\begin{eqnarray}
\label{eq:pdf-cut-eq}
\Pr(classification;T)\propto e^{-E(classification)/T}
\vspace{-1cm}
\end{eqnarray}
where the ``temperature'' $T$ serves as a free parameter, and
the energy $E$ takes into account both  unlabelled and labelled points.
Classification is then performed by marginalizing (\ref{eq:pdf-cut-eq}),
thus estimating the probability that a point $i$ belongs to a class $c$.
This formalism is often referred to  as a Markov random field (MRF),
which has been applied in numerous works, including in the context of semi-supervised learning
by~\cite{zhu03}. However,  they seek  the MAP solution (which corresponds to $T=0$),
while we estimate the distribution itself (at $T\ge 0$).

Using the framework of statistical physics has several advantages in the context of
semi-supervised learning:
First, classification  has a simple probabilistic interpretation.
It yields a fuzzy assignment of points
to class types, which may also  serve as a confidence level in the classification.
Secondly, since exactly estimating the marginal probabilities is, in most cases,  intractable,
 statistical physics has developed elegant Markov chain Monte-Carlo (MCMC) methods which are
suitable for estimating semi-supervised systems.
Due to the inherent complexity of semi-supervised problems, `standard' MCMC methods, such
as the Metropolis~\cite{Metropolis} and Swendsen-Wang~\cite{wang90} methods  provide poor results,
and one needs to apply more sophisticated algorithms, as discussed in section~\ref{sec:estimatingProb}.
Thirdly, using statistical physics allows us to gain an intuition  regarding  the nature of a semi-supervised
problem, i.e., it allows for a detailed analysis of the effect of adding labelled points to an
unlabelled data set.
In addition, our method also has two practical advantages:
(i) while most semi-supervised learning methods consider only the case of
two class types, our method is naturally extended  to  the multi-class scenario.
(ii) Another unique feature of our method is its ability to suggest
the existence of a new class type, which did  not appear in the labelled set.

Our main objective in this paper is to present a framework,
which can later be applied in  different directions.
For example, the energy function in (\ref{eq:pdf-cut-eq}) can be any of the functions
used in other semi-supervised  methods. In this paper we chose to use the min-cut cost function.
We do not claim that using this cost function is optimal, and indeed we observed
that it is suboptimal in some cases.
However, we aim to convince the reader that applying our method, to any energy function,
would always yield equal or better results than merely minimizing the same energy function.
\newpage
Our work is closely related to the typical cut criterion
for unsupervised learning,
first introduced by~\cite{blatt96} in the framework of statistical physics
and later in a graph theoretic context   by~\cite{GdalyahuEtAl99}.
The method introduced in this work can be viewed as  an
extension of these clustering algorithms to the semi-supervised case.

The paper is organized as follows:
Section~\ref{sec:modelDef} presents the model, and Section~\ref{sec:estimatingProb} discusses
the issue of estimating  marginal probabilities.
Section~\ref{sec:classification} presents the qualitative effect of adding labelled points.
Our semi-supervised algorithm is outlined in Section~\ref{sec:algorithm}.
Section~\ref{sec:results} demonstrates the performance of our algorithm
on a toy data set  and on two real-life examples  of  gene expression data.

\vspace{-0.25cm}
\section{Model definition}
\label{sec:modelDef}
\vspace{-0.15cm}
In our model each data point $i, i=1,\ldots, N$, corresponds to a  random variable, or  spin,
$s_{i}$ which can take one of $q\geq 2$ discrete states.
The number of states $q$ matches the number of class types in the labelled set.
A certain classification of the data set then corresponds to a vector ${\bf S}$,
${\bf S}=\{s_{1},\ldots, s_{N}\}$.
Assume that the first $M\ll N$ points are labelled, i.e.,
the state of spin $s_{k}, 1\le k\le M$   is  clamped to a  spin
value $c_{k}$, which   corresponds to the class type of point $k$.
Hence the energy $E$ in our case is simply the Potts model energy of
a {\em granular ferromagnet} with an external field;
\begin{eqnarray}
\label{eq:Hamiltonian}
E({\bf S})=\sum_{\langle i,j\rangle} J_{ij}(1-\delta_{s_{i},s_{j}}) + \sum_{k=1}^{M} h_{k}(1-\delta_{s_{k},c_{k}}).
\end{eqnarray}
where $J_{ij}>0$ is a predefined similarity between points $i$ and $j$,
$\langle i,j\rangle$ stands for all edge of neighboring graph,
and $\delta_{s_{i},s_{j}}=1$ when $s_{i}=s_{j}$ and zero otherwise.
The second term which corresponds to the labelled points, is known as the `external field' term.
In case the value of $s_{k}$ is different from the point's assigned class $c_{k}$, the energy
is increased by a value $h_{k}$.
We used   $h_{k}=\infty$,  which assigns
non-zero probability only to classifications in which  $s_{k}=c_{k}$.
Notice that one can introduce uncertain labels by using finite values  of $h_{k}$,
but we do not consider this case in this work.

The major problem in applying the suggested method concerns the difficulty in calculating
(\ref{eq:pdf-cut-eq}). Since the number of possible classifications is exponential in $N$,
one often needs to apply sampling MCMC algorithms, which are considered in the next section.

\section{Estimating marginal probabilities}
\label{sec:estimatingProb}
Introducing labelled points inherently changes the properties
of the system and poses great difficulties in MCMC sampling.
Labelled points may introduce  `frustration' into the system
(a term borrowed from statistical physics); if, for example, point $i$ is connected
to a couple of differently labelled points $j$ and $k$, it is `frustrated' since
whenever it matches one of them it contradicts the other.
Such frustration appears also in physical systems of {\em spin glasses},
and is known to complicate their analysis.

The difficulty in sampling from spin glass systems results from their
ragged energy landscape.
The energy landscape can be described as being composed of
several `valleys' which are surrounded by very high energy barriers,
which the sampling method is unable to traverse
at low temperatures.
As a results, `standard' MCMC methods, e.g. the Metropolis and the Swendsen-Wang methods,
are confined to a certain `valley' for an exponential number of Markov chain steps,
 thus  their  estimates may be  highly biased.

{\em Extended MCMC} methods is a title given for a family of methods which
enable efficient sampling in complex scenarios such as spin-glasses~\cite{iba01}.
Extended MCMC methods solve the sampling  problem by allowing the system to
`jump' between `valleys'. This is implicitly performed by letting
the system pass through high energy configurations, which most
likely erase any memory of the originating `valley'.
In this work we applied the {\em Multicanonical} Monte-Carlo method~\cite{berg92a},
which is a member of the extended MCMC methods.

The Multicanonical Monte-Carlo method first estimates the density of states $D(E)$,
i.e. the number of different classifications  at a given energy.
It then generates a sample of classifications, $\{{\bf S}\}$, drawn from the
distribution $Pr_{D}({\bf S})\propto 1/D(E({\bf S}))$,
which can  then be used   to recover the
Boltzmann distribution (\ref{eq:pdf-cut-eq}), for all temperatures $T$ at once.
Sampling from   $Pr_{D}({\bf S})$ yields a uniform  distribution
over     all   energy levels, which forces the MCMC
to pass through high energy configurations and by that overcome the energy barriers.
For further  details  about the method we refer the reader to~\cite{iba01}.

Before presenting  the effect of labelled points,
we would like to shortly discuss an alternative to MCMC sampling, which is
 to approximate the marginal probabilities  using methods from the field of graphical models.
The intimate  connections between statistical physics and graphical models
have been demonstrated, e.g. by~\cite{YedidiaEtAlIJCAI01}. Our Boltzmann distribution corresponds
to an undirected graphical model, thus estimating the marginal probabilities is equivalent
to performing inference in this model. Since exact inference, via the
junction tree algorithm~\cite{Pearl88}, is generally intractable, one needs to resort to
approximate methods, such as (loopy) belief propagation (BP)
or generalized belief propagation (GBP)~\cite{YedidiaEtAlIJCAI01}.
In our experimental study the performance of BP  was rather poor,
probably since the graphical model of a typical semi-supervised problem contains many short loops.
On the other hand the performance of GBP was excellent, when applied to two dimensional
problems, as in Fig.~\ref{fig:toy}.
However, to date there exists no principled way of applying GBP to a general graph,
which guarantees  good approximate inference, therefore we consider only MCMC sampling.

\section{The effect of labelled points}
\label{sec:classification}
As explained in section~\ref{sec:estimatingProb} the  labelled points
inherently complicate the sampling of the system.
On the other hand, adding labelled points has the desired effect on classification.
In order to understand this phenomenon we first describe
the unsupervised case and then qualitatively explain the effects of adding labelled points.

The properties of a system, governed by the Boltzmann distribution (\ref{eq:pdf-cut-eq}),
changes with the temperature $T$.
In many physical systems, the temperature range $T\geq 0$ can be divided
into intervals, or {\em phases}, each of which has its own global
properties. Granular-ferromagnets without  external fields
(i.e. keeping the first term of (\ref{eq:Hamiltonian})),
which correspond to the unsupervised case, are known to have three phases~\cite{wiseman98};
a low-temperature phase in which the system is ferromagnetic,
i.e. most of the spins are assigned the same value;
a high temperature phase in which the system is paramagnetic, i.e.
the values assigned to the spins are nearly independent;
and an intermediate phase termed the super-paramagnetic (SP) phase.
In this phase, which is the most relevant for clustering,
all spins of a  grain (i.e. a cluster) are assigned a certain value, with
different values at different grains.
The clusters in the data can be identified in this SP phase;
the larger the temperature interval of this phase,
the more significant and stable is the clustering solution~\cite{levine}.

Adding labelled points changes the system's behavior.
First, it effectively increases the strength of the interaction
between  spins near labelled points, which can be interpreted
 as an increase of their local density.
As a result there is an increase in the transition temperature between the ordered SP phase
and the unordered paramagnetic phase, thus increasing  temperature
interval of the SP phase at its the upper limit.

A second effect happens at low temperatures. For example, consider the case of two dense grains,
each containing a labelled point of a different type, which are separated by a lower density region.
In the SP phase the spins in each of the grains attain their correct class, but the spins in the low density region
are still unordered. As the temperature is  lowered the two classes `penetrate'
into the low density region until a `border' between the classes is formed.
Hence, from a semi-supervised perspective,
the labelled points cause the low density region to be classified.
Notice that at this temperature the unsupervised case is already at
the ferromagnetic phase, where the two clusters are united.
Hence, the labelled points also decrease the lower limit of the SP  phase,
which together with increasing its upper limit, results in a larger temperature interval
relevant for classification.

When the temperature is further lowered,
a different classification may appear. For example, one of the class types may
overtake the whole system, similar to the min-cut solution in Fig.~\ref{fig:toy}d,
but, of course,  we are not interested is such a solution.

Fig. 2 presents the effect of adding labelled points
in the case of Fig.~\ref{fig:toy}a.
We plot the number of misclassified points using the algorithm in Sec.~\ref{sec:algorithm},
 as a function of $T$, in the unsupervised (US) and in the semi-supervised (SS) cases.
In this data set we calculated (\ref{eq:pdf-cut-eq}) exactly using the junction
tree algorithm (exact US and exact SS), and compared it to Multicanonical sampling (MC).
Notice that  adding labelled points decreases the number of errors
dramatically, achieving almost correct classification over a large temperature interval ($0.5 \le T \le 1$).
At lower temperatures, which correspond to the min-cut solution (Fig.~\ref{fig:toy}d), the number of misclassified
points is large.
\vspace{-0.3cm}
\begin{figure}[thb!]
\parbox{4.75cm}{
\begin{tabular}{c}
\psfig{file=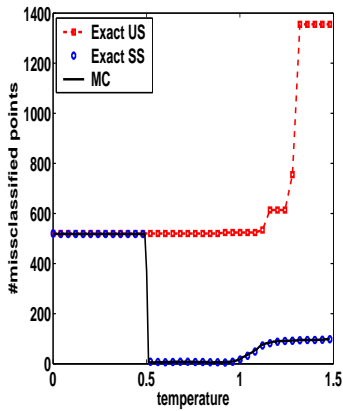,width=4.5cm,height=5.5cm}
\end{tabular}
} \hfill
\parbox{3.25cm}
{
\caption{\capfont Classification error as a function of $T$ for the data set
in Fig.\ref{fig:toy}a. A comparison between the unsupervised case
(exact US) and the semi-supervised case (exact SS),
calculated exactly using the junction tree algorithm,
and the Multicanonical MCMC method (MC). }
}
\label{fig:toyErrors}
\vspace{-0.5cm}
\end{figure}

\vspace{-0.2cm}
\section{The algorithm}
\label{sec:algorithm}
\vspace{-0.2cm}
Our semi-supervised learning algorithm is comprised of two parts:
an estimation part, and a classification part which are described below.
\vspace{0.3cm}
\hrule
\vspace{-0.1cm}
\paragraph{Estimation}\hspace{-0.2cm}consists of three stages:
\newline $\bullet$ Map each point $i$ to a $q$-state random variable $s_{i}$, where $q$
is the number of class types of labelled points.
\newline $\bullet$ Construct a graph $G$ of neighboring points $i$ and $j$ and
assign a their pairwise similarity $J_{ij}>0$.
\newline $\bullet$ Estimate the marginal probabilities $p_{i}(s_{i};T)$ and $p_{ij}(s_{i},s_{j};T)$,
as explained in Sec.~\ref{sec:estimatingProb}.

\paragraph{Classification}\hspace{-0.2cm} of  a point $i$ at a temperature $T$
can simply be performed by $\mbox{argmax}_{1\le k \le q} p_{i}(s_{i}=k;T)$.
However, we suggest a heuristic method which is slightly more elaborate, but takes
into account the confidence in the classification, and also allows to identify
new class types. This heuristic is comprised of two steps:
\newline $\bullet$ {\em Classify `confident' points}, using single point probabilities $p_{i}$:
For each point $i$ we find the {\em two} most probable class assignments
and their probabilities; $p_1$ and $p_2$. In case $p_{1}-p_{2}>\tau$,
where $\tau>0$ is a user defined confidence parameter,
we classify point $i$ according to the type which corresponds to $p_1$.
\newline $\bullet$ {\em Classify  the remaining,  less `confident' points},
using the pairwise probabilities $p_{ij}$:
Following the intuition of~\cite{blatt96} we estimate the pairwise
correlations between $i$ and $j$ defined as
\vspace{-0.15cm}
\begin{eqnarray}
\nonumber
C_{ij}(T)=\sum_{k=1}^{q}p_{ij}(s_{i}=k,s_{j}=k;T),
\vspace{-0.15cm}
\end{eqnarray}
where the correlation ranges from the random level $1/q$, to a perfect correlation value of $1$.
We then delete edges from the  graph G for which $C_{ij}(T)<\frac{1}{2} \left( 1 + \frac{1}{q} \right)$,
i.e., half way between random level and perfect correlation,
and find  the connected components of the resulting graph.
Each `unconfident' point $j$ is then classified according to the connected component to
which it belongs.
In case $j$ belongs to a connected component which contains points (already)
classified as $c_{k}$, then  $j$ is assigned to $c_{k}$.
If $j$ belongs to  a connected component which contains points which were
assigned to several different classes, it   is remains unassigned and is marked
as ``confused" between these classes.
Finally, all the points which belong to a connected component that
does not contain any classified point are marked as a new class.
\vspace{0.3cm}
\hrule
\vspace{0.1cm}

Notice that the classification depends on $T$.
The rational is to supply the user with a classification `profile' of each data point,
over all temperatures.
Since statistically significant classifications span large temperature intervals,
such a `profile' is rather limited in size.
For example, the `profile' of a point $j$ which resides in class $c_{1}$,
close to the border with class $c_{2}$ would contain two classifications:
at low temperatures $j$ is assigned to $c_{1}$, and at higher temperatures it is marked
as ``confused" between $c_{1}$ and $c_{2}$.
\newline
As for the value of $\tau$,  our experimental study has shown that
classification  performance decreases with increasing the value of $\tau$ (data not shown),
thus we chose to use  $\tau=0.1$.

In case there are no labelled points $p_{i}(s_{i}=k)=1/q ~\forall i,k$ then
all points are treated as `unconfident', and our `classification' simply coincides
with the clustering procedure of~\cite{blatt96}.

\section{Experimental results}
\label{sec:results}
We present results over three data sets: A toy data set, and two
real-life data sets of gene expression.
\subsection{Toy data}
\label{sec:expResToyData}
Fig.~3 presents a  toy data set similar to Fig.~\ref{fig:toy},
which contains $1306$ data points from three classes.
As in the former toy data, the  similarity between
adjacent pixels is of unit value, hence  the three  classes form
one connected cluster, which can not be separated without the labelled points.
\vspace{-0.2cm}
\begin{figure}[thb!]
\parbox{4cm}{
\begin{tabular}{c}
\psfig{file=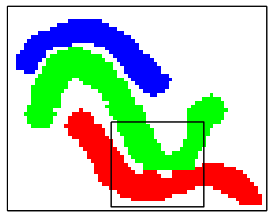,width=3.5cm}
\end{tabular}
} \hfill
\parbox{4cm}
{
\caption{\capfont A toy data set comprised of three classes.
Each pixel corresponds to a data point, and $J_{ij}=1$ for all adjacent pixels $i$ and $j$.
The labelled points are randomly sampled with a uniform distribution.
In order to  enable correct classification the labelled points
from the lower two classes are sampled  from the area marked by a  rectangle.
}
}
\label{fig:threeWaves}
\end{figure}

\begin{figure*}[thb!]
\begin{center}
\begin{tabular}{cccc}
\psfig{file=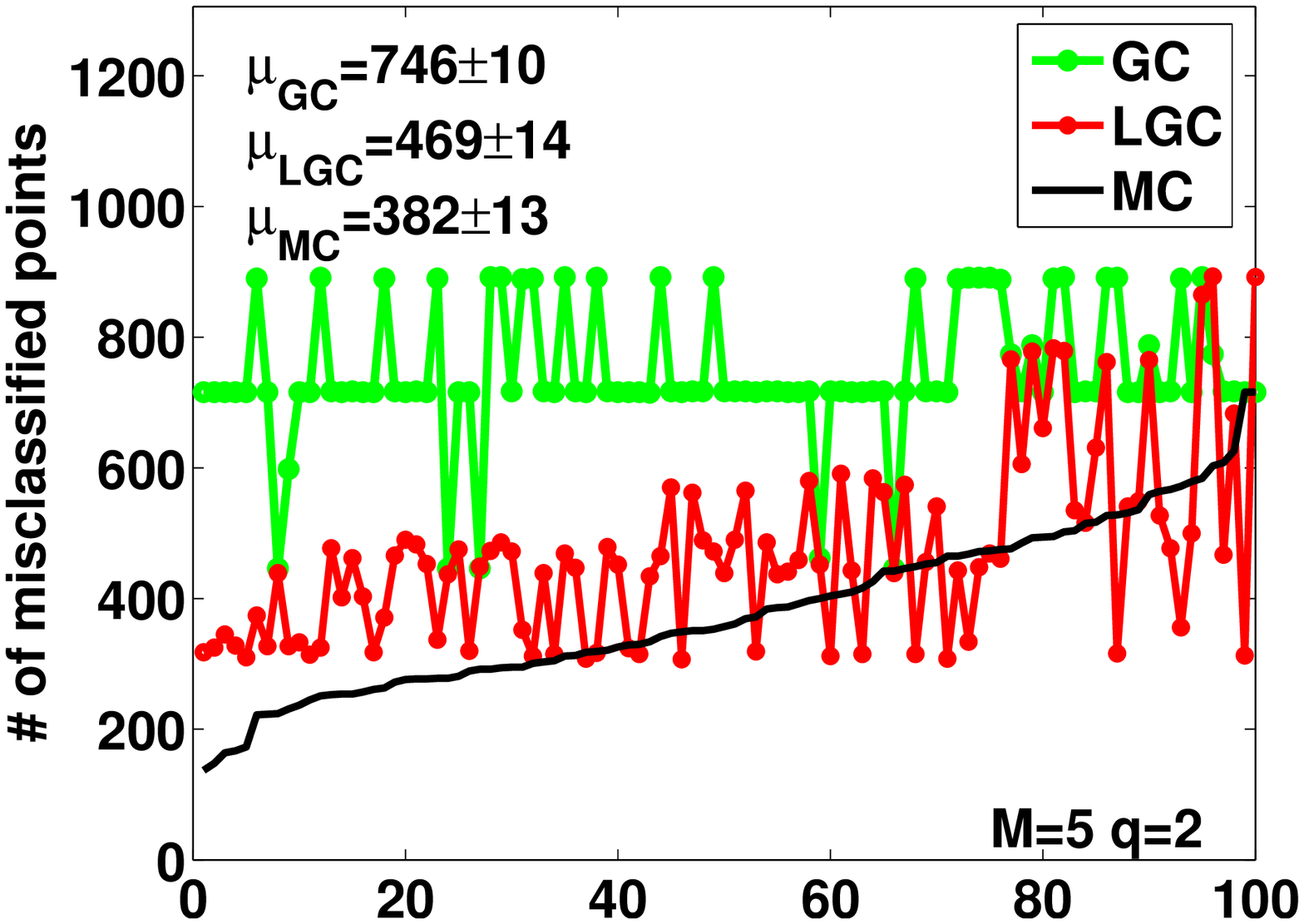,width=4.2cm}&
\hspace{-0.5cm}
\psfig{file=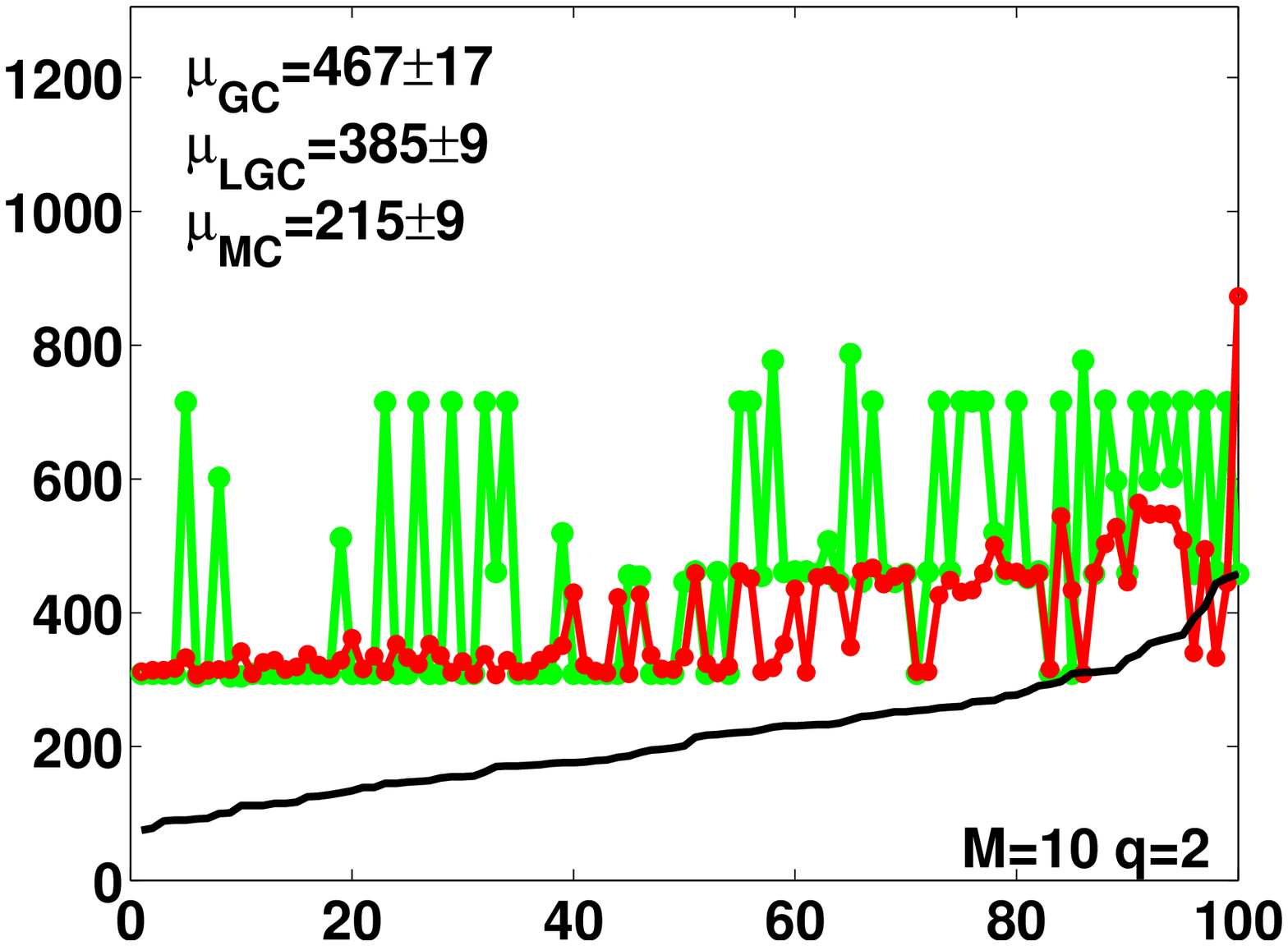,width=4cm}&
\hspace{-0.5cm}
\psfig{file=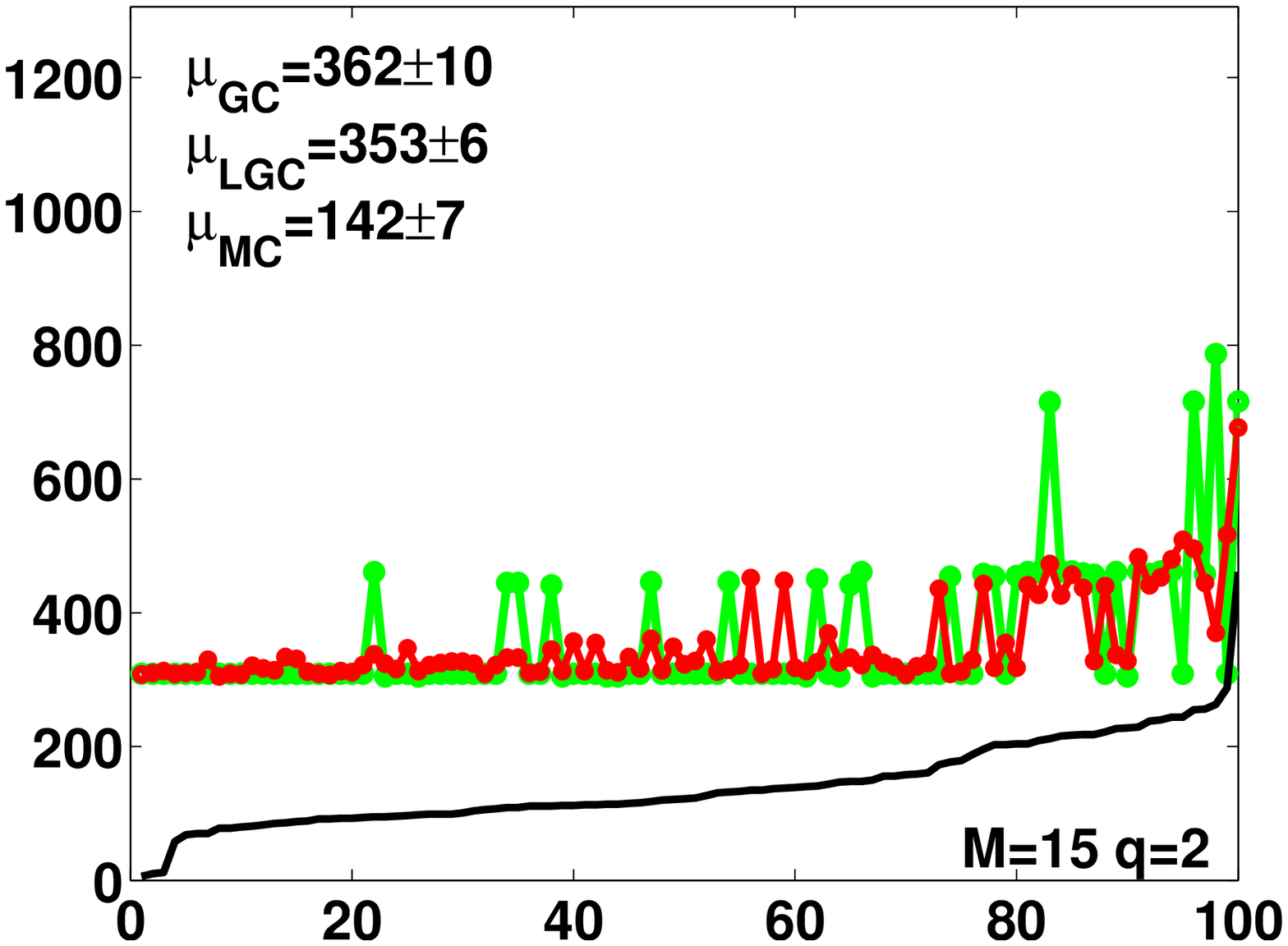,width=4cm}&
\hspace{-0.5cm}
\psfig{file=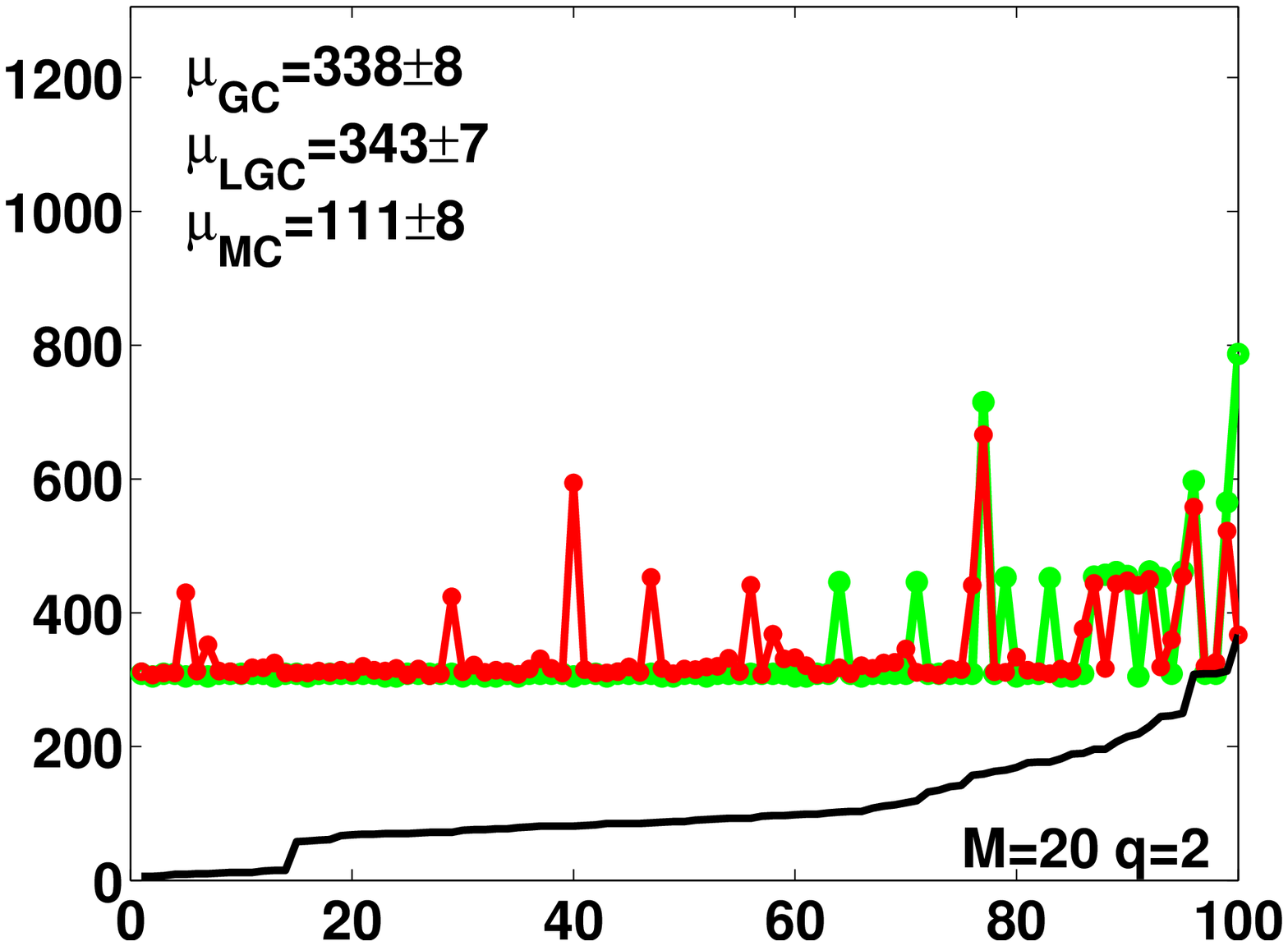,width=4cm}\\

\psfig{file=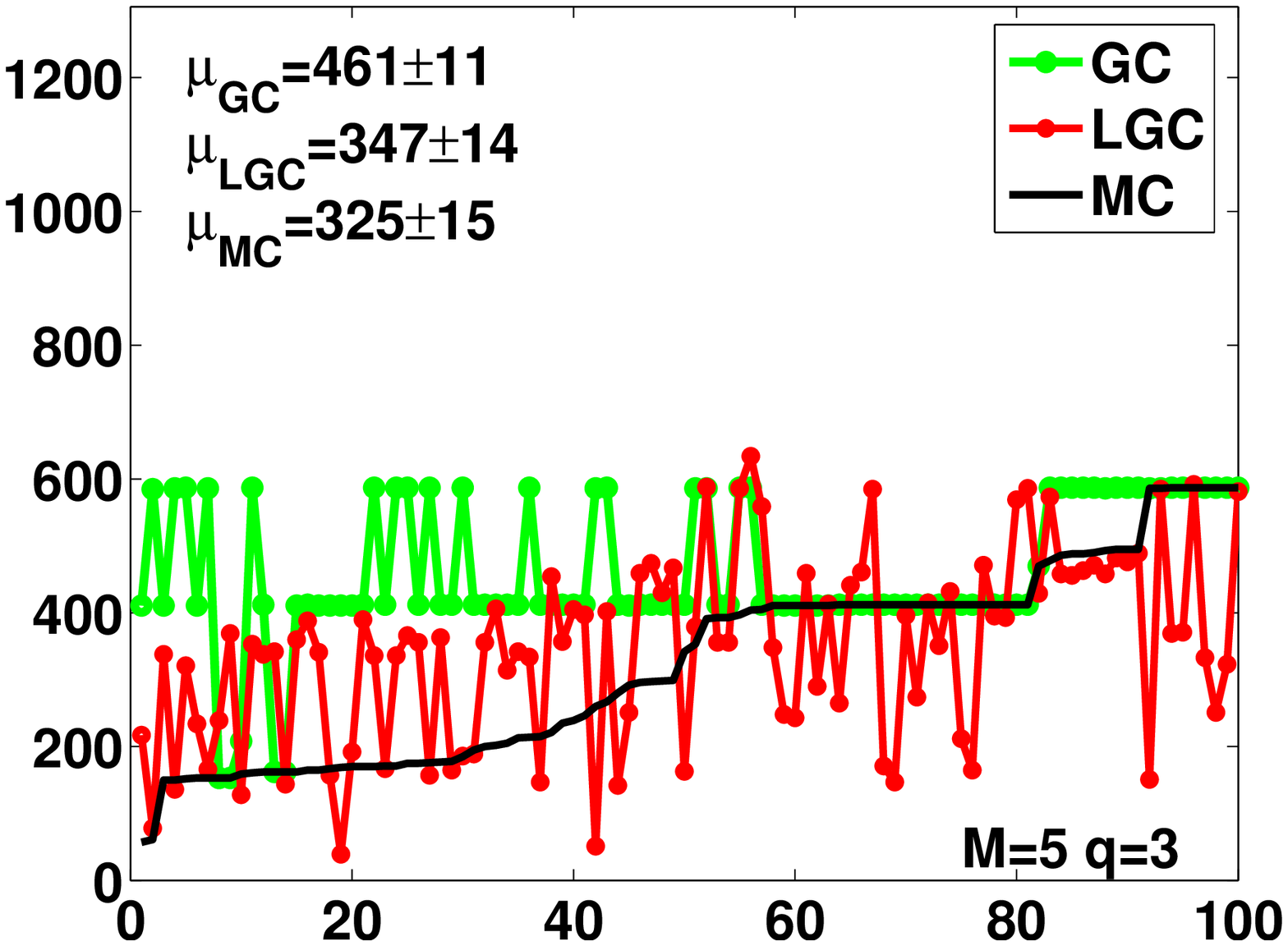,width=4cm}&
\hspace{-0.5cm}
\psfig{file=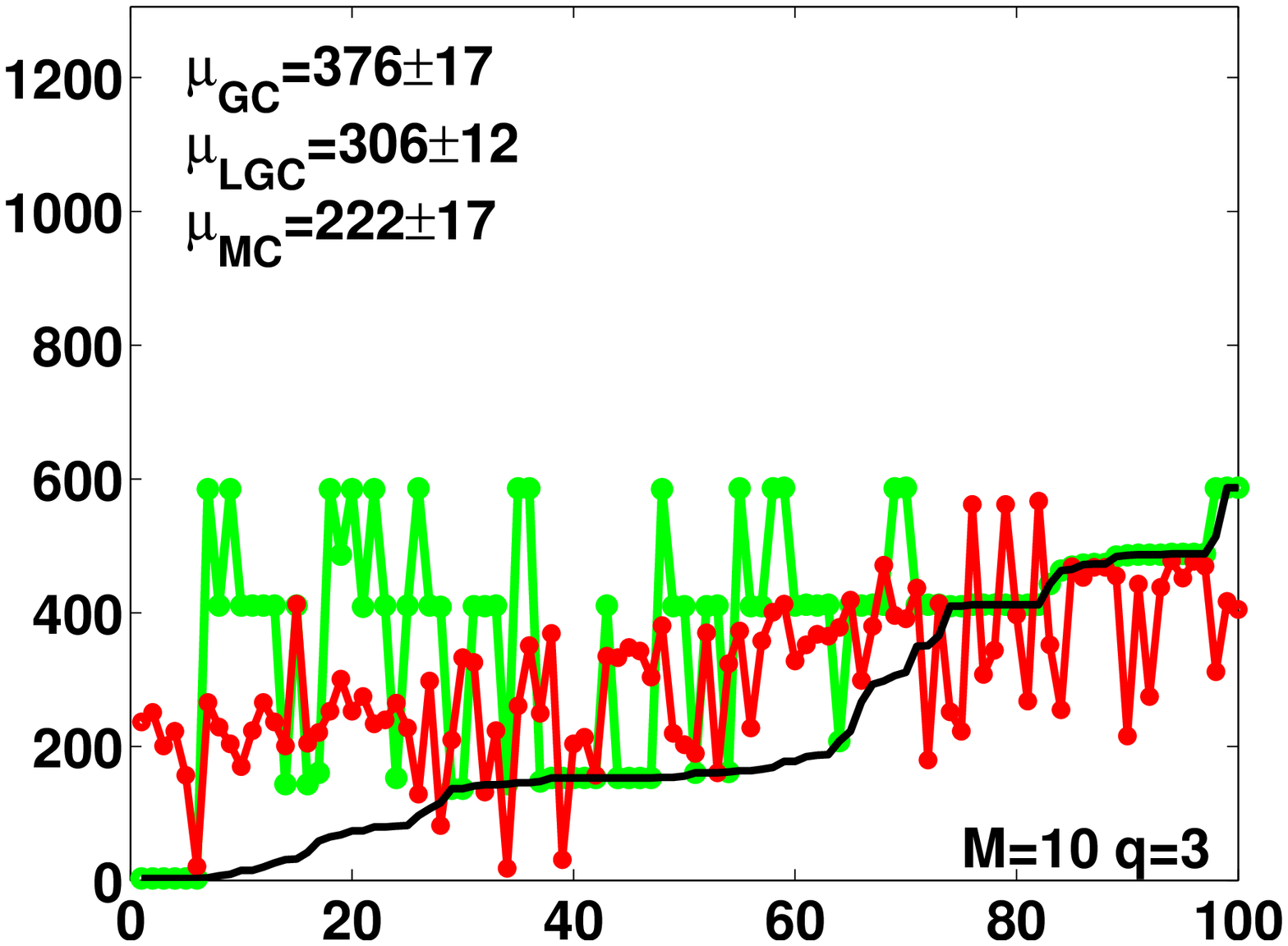,width=4cm}&
\hspace{-0.5cm}
\psfig{file=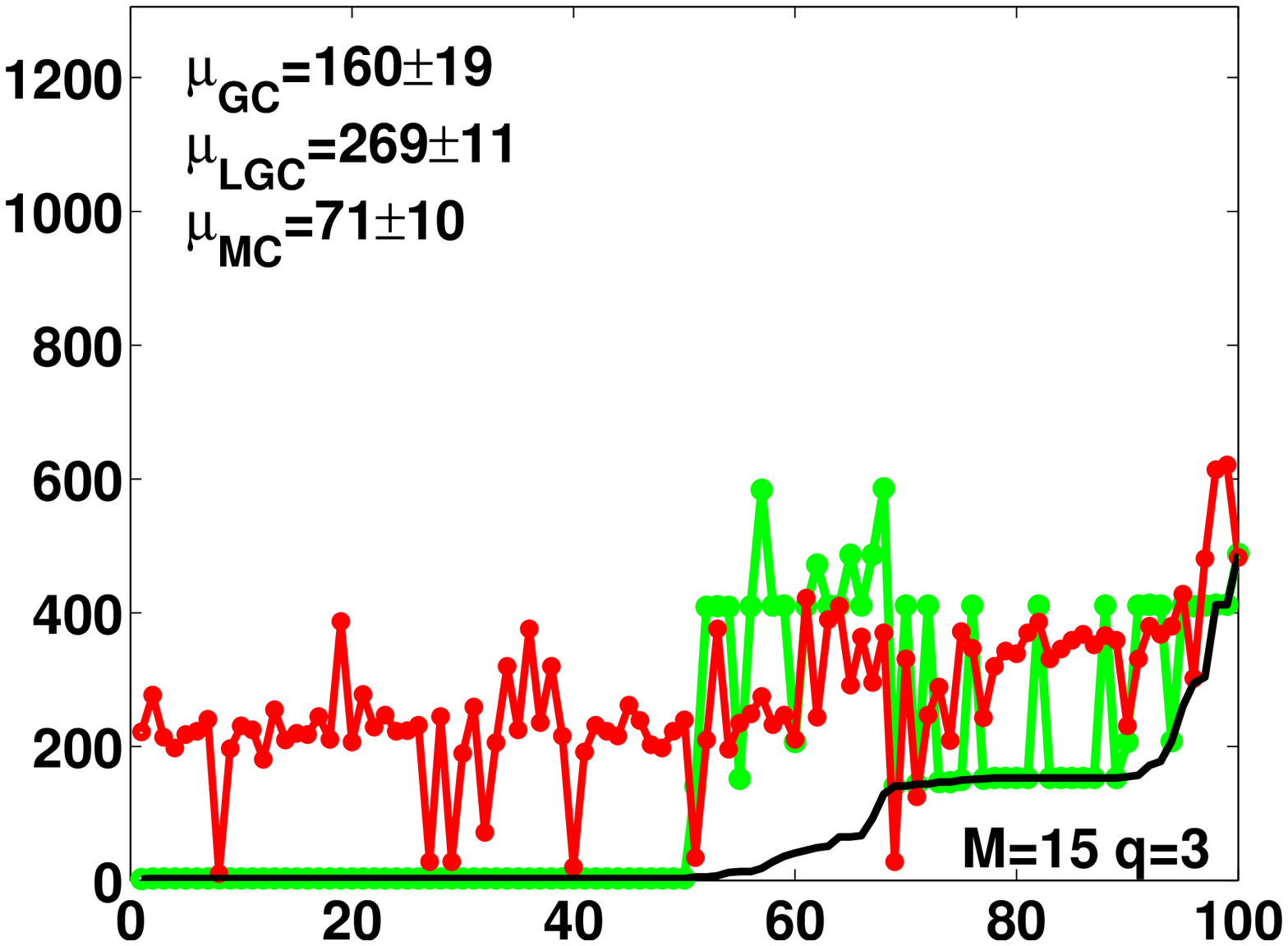,width=4cm}&
\hspace{-0.5cm}
\psfig{file=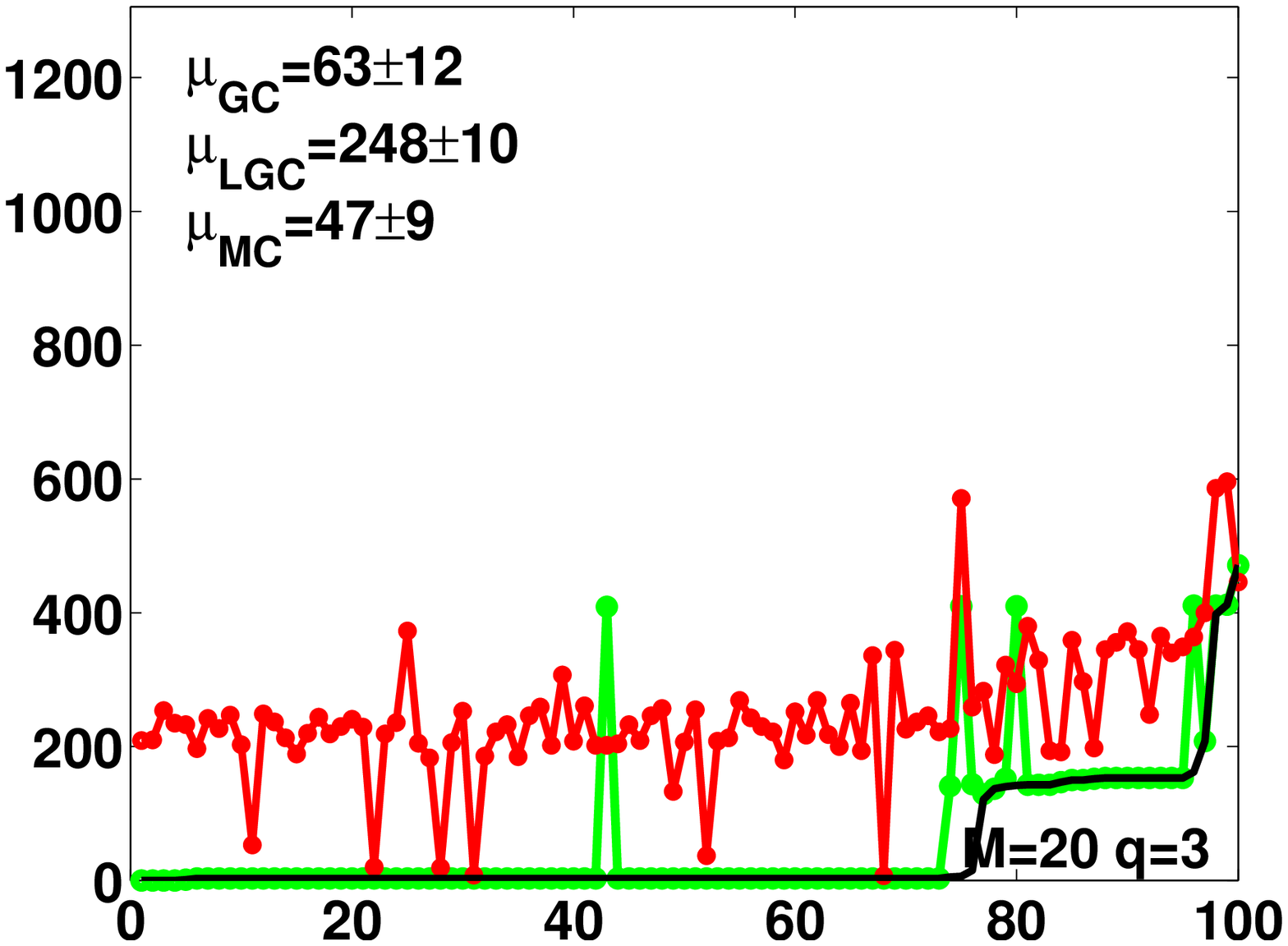,width=4cm}\\
\hspace{-0.5cm}
\end{tabular}
\end{center}
\vspace{-0.75cm}
\caption{\capfont The number of misclassification out of $1306$ points of Fig.~3,
for $q=2$ (upper row) and for $q=3$ (lower row) for $100$ random samples
of labelled points $M$ ($M=5,10,15$ and $20$ from left to right).
We compare our method (MC, black line) to graph-cuts (GC, red line) and to
the local-global method (LGC, green line).
Also shown is the mean performance $\mu$ and its standard deviation.
In the unsupervised case the number of misclassified points was $1001$.}
\label{fig:threeWavesClassification}
\vspace{-0.5cm}
\end{figure*}

In order to evaluate the performance of our approach we carried two
sets of experiments. In the first set we randomly chose
$M$\footnote{$M$ denotes the total number labelled points.}
labelled points ($M=5,10,15$ and $20$) from
the two lower (green and red) classes, i.e. $q=2$, while in
the second set of experiments the labelled points where randomly chosen from all three
classes, i.e. $q=3$.
For each value of $M$ and $q$ we evaluated $100$  instances (realizations)
of random labelling, and the number of misclassified points
appear in Fig.~\ref{fig:threeWavesClassification}.
The  number of misclassified points in  the unsupervised case was $1001$,
and was evaluated as explained in Sec.~\ref{sec:algorithm}.

As expected, incorporating even a few labelled points has a significant
impact on the number of misclassified points.
As can be seen, the results  highly depend on the specific instance of
labelled points, hence the average performance is less informative.
Therefore the instances are presented in an increasing order of misclassified points of
our approach (MC), while the other two lines
correspond to the graph-cuts method\footnote{In order to apply graph-cuts  when
$q=3$ we used the approximation of~\cite{boykov99}.} (GC)
and to the local-global consistency method~\cite{scholkopf}  (LGC).
In order to plot the MC line in Fig.~\ref{fig:threeWavesClassification}
we automatically selected a temperature, $T^{*}$,
in which classification is significantly different
from the ground state ($T=0$) solution, and is also most `stable'.
At each $T$ we consider only the points, $c(T)$,  whose
classification is both confident and different than the $T=0$ solution.
We  define a score $\eta(T)=\| c(T)\| \cdot  s(T)$,
 where $s(T)$ is the average temperature interval in which the classifications
 of $c(T)$ remain unchanged.
Then, $T^{*}=\arg \max \eta(T)$, and in case  $\eta(T^{*})<\eta_{0}$, we set
$T^{*}=0$. In general,  we recommend to use the `profiles' of the points,
since there may be several `stable' solutions at different temperatures.

In comparing our method and  graph-cuts, both of which use the same energy function,
it can be observed that our method  always  achieves an equal or lower number of
misclassifications than graph-cuts.
However, it appears that in  several  instances of labelled points,
it is preferable to apply the  energy function of~\cite{scholkopf}.
\newline
Also it seems that for $q=2$ our method significantly outperforms
the other two methods, mainly due to its ability to identify the third class type,
although none of its points is labelled.
For $q=3$ the solutions of graph-cuts and of
our method become similar as the number of labelled points increases.

\vspace{-0.2cm}
\subsection{Leukemia gene expression data set}
\label{sec:leukemia}
\vspace{-0.2cm}
In this section we present the results of
applying our algorithm to a real-world problem of cancer
classification and class discovery. In cancer research,
there is a particular need for semi-supervised techniques,
as the classes  and sub-classes (cancer types) are
only partially known.
Hence one needs to apply methods that can help partition the data into
known classes and possibly identify novel ones.

Our example is based on gene expression data\footnote{Simultaneous measurements
of mRNA levels of thousands of genes in a single tissue sample.}
of acute leukemia published by~\cite{armstrong02}.
They analyzed three different types of acute
leukemia; acute myeloid leukemia (AML), acute lymphoblastic
leukemia (ALL) and a sub-type of ALL which carries a chromosomal
translocation in the MLL gene. Armstrong et al.~ show (in a
supervised manner) that this sub-type  has a distinct molecular
profile and can be considered a new type of leukemia termed  MLL.

We applied our algorithm to the 57 leukemia samples in~\cite{armstrong02}
($20$ ALL, $20$ AML and $17$ MLL samples),
each described by the expression levels
of the 200 genes with largest variance across samples.
The similarity between samples was calculated in a standard manner
in this field\footnote{The expression level of each gene is `normalized' by subtracting
its mean expression over all samples, and divided by its standard deviation.
The distance between samples $i$ and $j$, $d_{ij}$, is then the Euclidean norm
over their $200$ genes, and $J_{ij}=\exp(-d_{ij}^2/a^{2})$ where $a=\langle d \rangle$ .}.
The same as in Sec.~\ref{sec:expResToyData}, we carried two set of experiments.
In the first set of experiments we randomly chose $M$  points ($M=2,4,6$)
from the ALL and AML samples but not from the MLL class, and in the second set of experiments
$M$ labelled points ($M=3,6,9$) were randomly selected from all three classes.
The results appear in Fig.~5 in the same format as in
Fig.~\ref{fig:threeWavesClassification}.
The  number of misclassified points in  the unsupervised case was $11$.

As in the previous data set, our method always
achieves an equal or lower number of misclassifications than graph-cuts.
Notice that in the $q=2$ case, our method is able to predict the existence of MLL,
while all  $17$ MLL points are misclassified in the other methods.
It appears that   for this data set, applying the min-cut
cost function is almost always superior to the quadratic cost function of~\cite{scholkopf}.
Another interesting phenomenon is the relatively low number of misclassifications
in the unsupervised case. It happens that in $20\%-40\%$ of the instances
(depending on $q$ and $M$) it is preferable to apply our method {\em without}
the labelled points.
\begin{figure*}[thb!]
\parbox{4cm}{
\begin{center}
\begin{tabular}{ccc}
\psfig{file=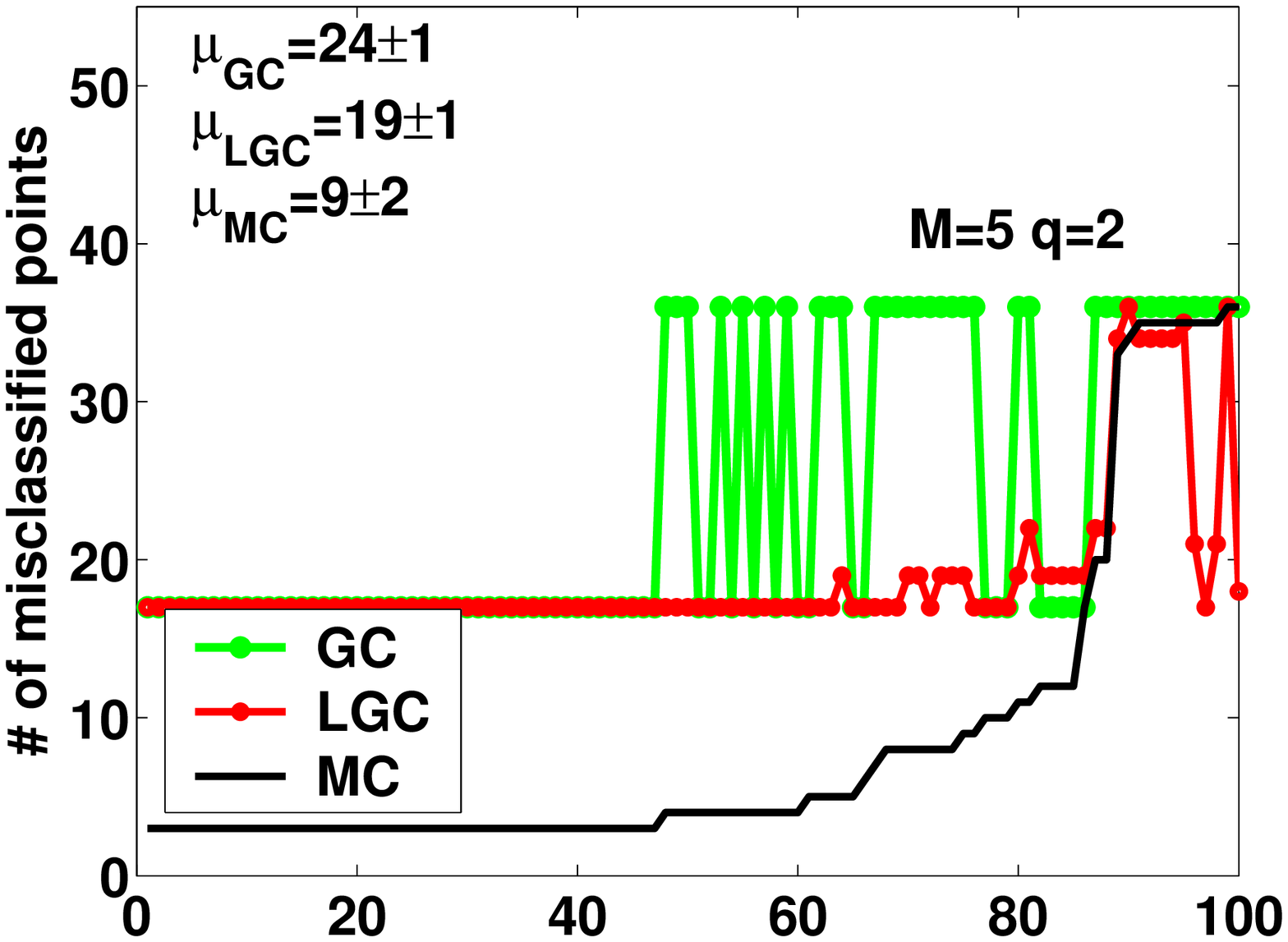,width=4.2cm}&
\hspace{-0.5cm}
\psfig{file=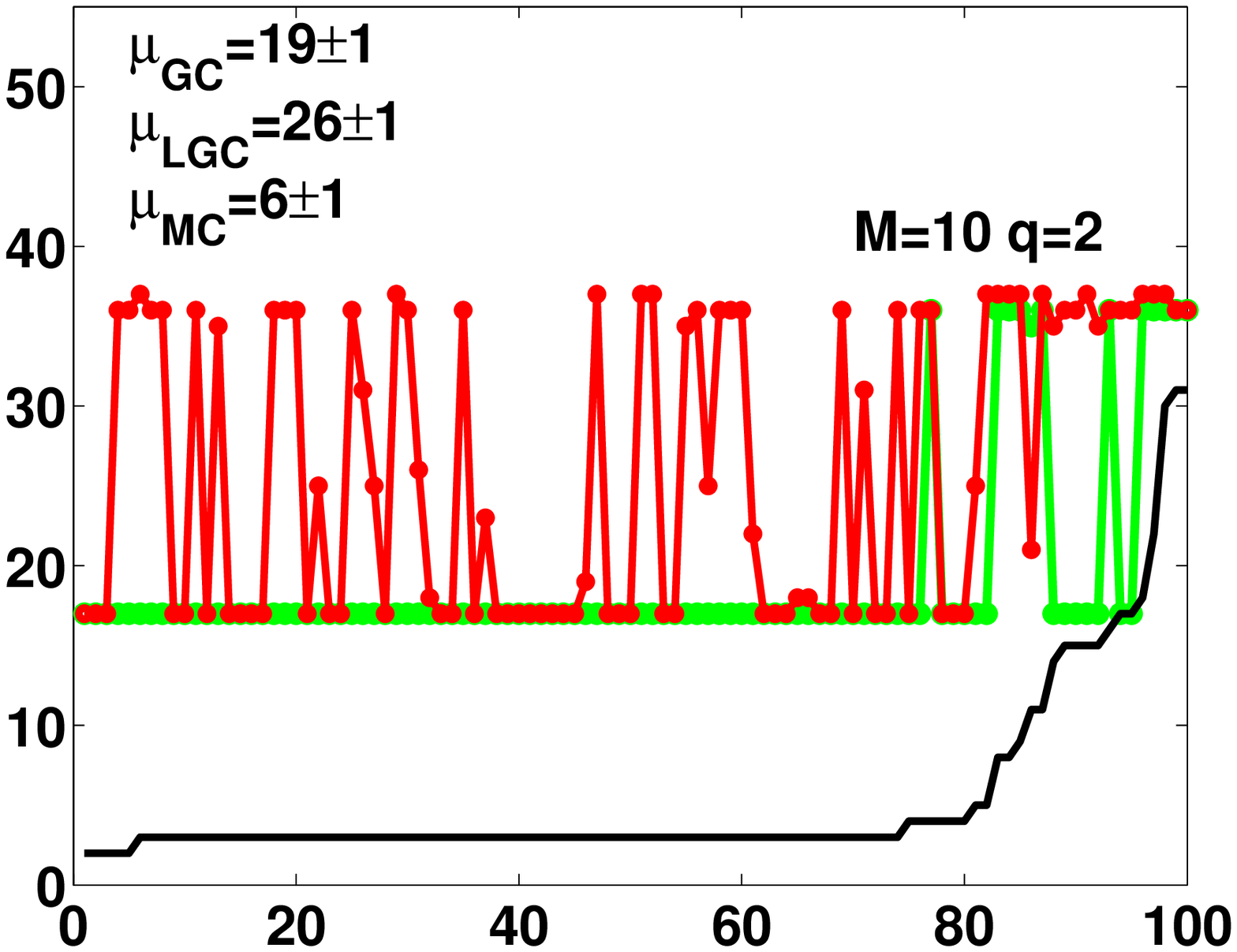,width=4cm}&
\hspace{-0.5cm}
\psfig{file=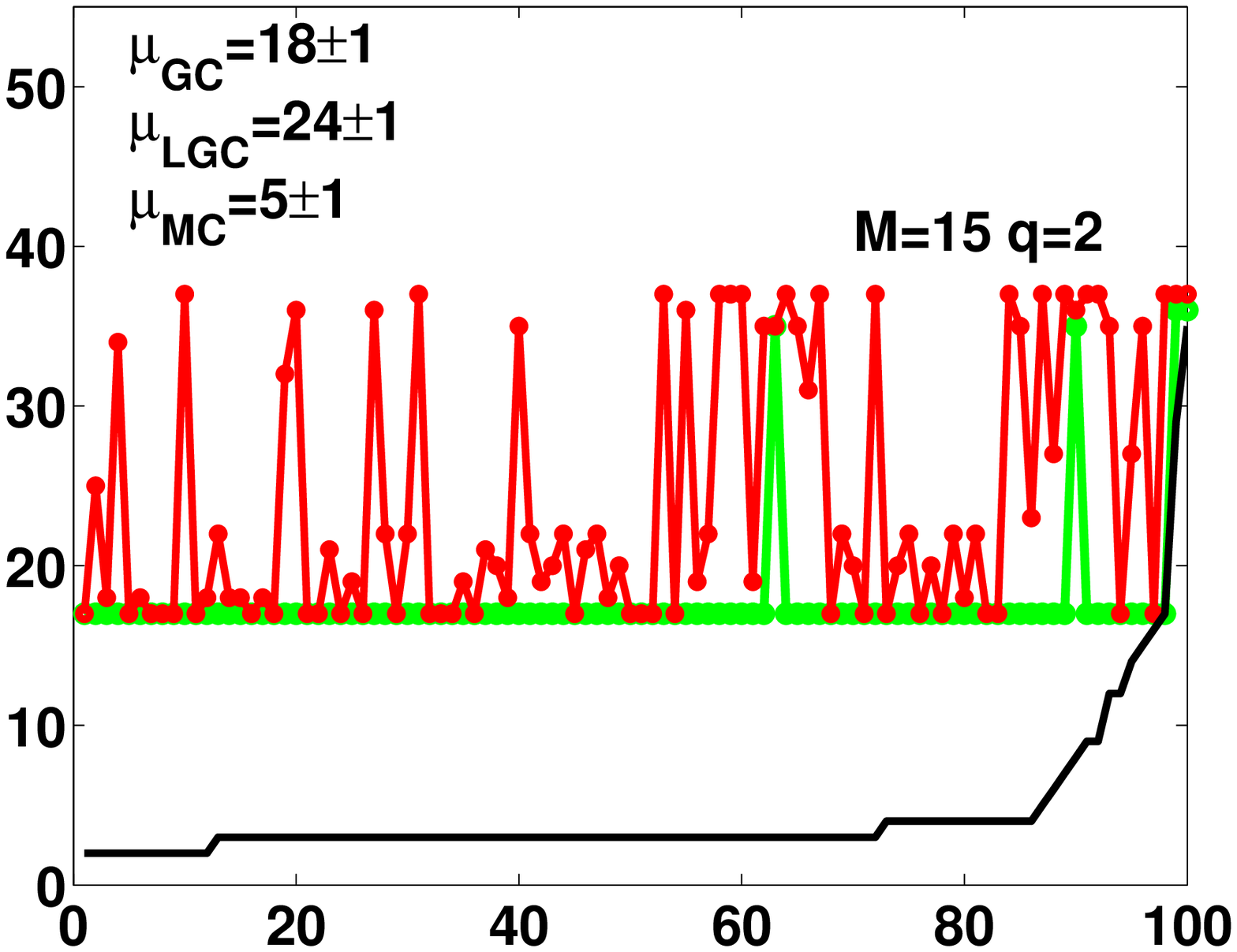,width=4cm}
\\
\psfig{file=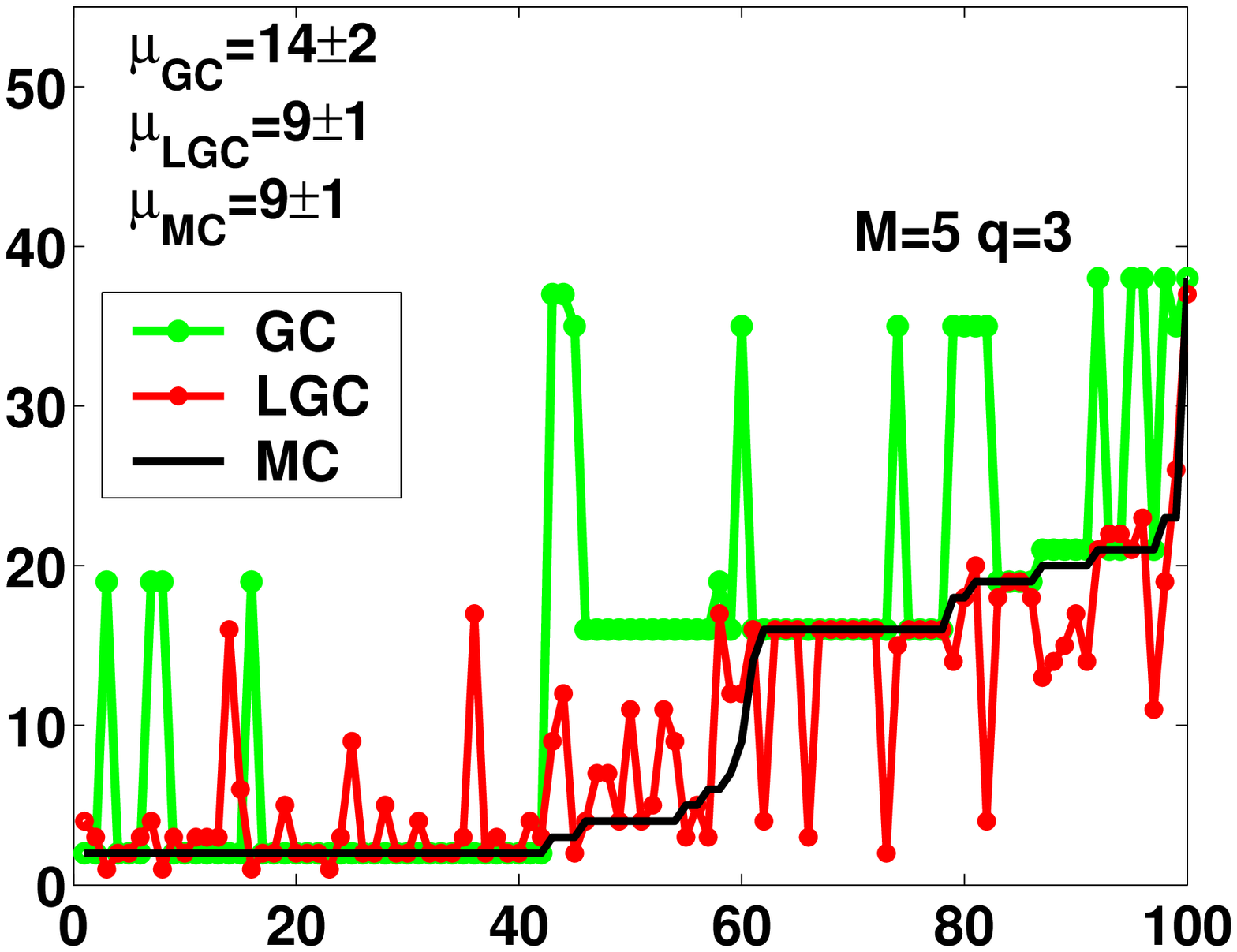,width=4cm}&
\hspace{-0.5cm}
\psfig{file=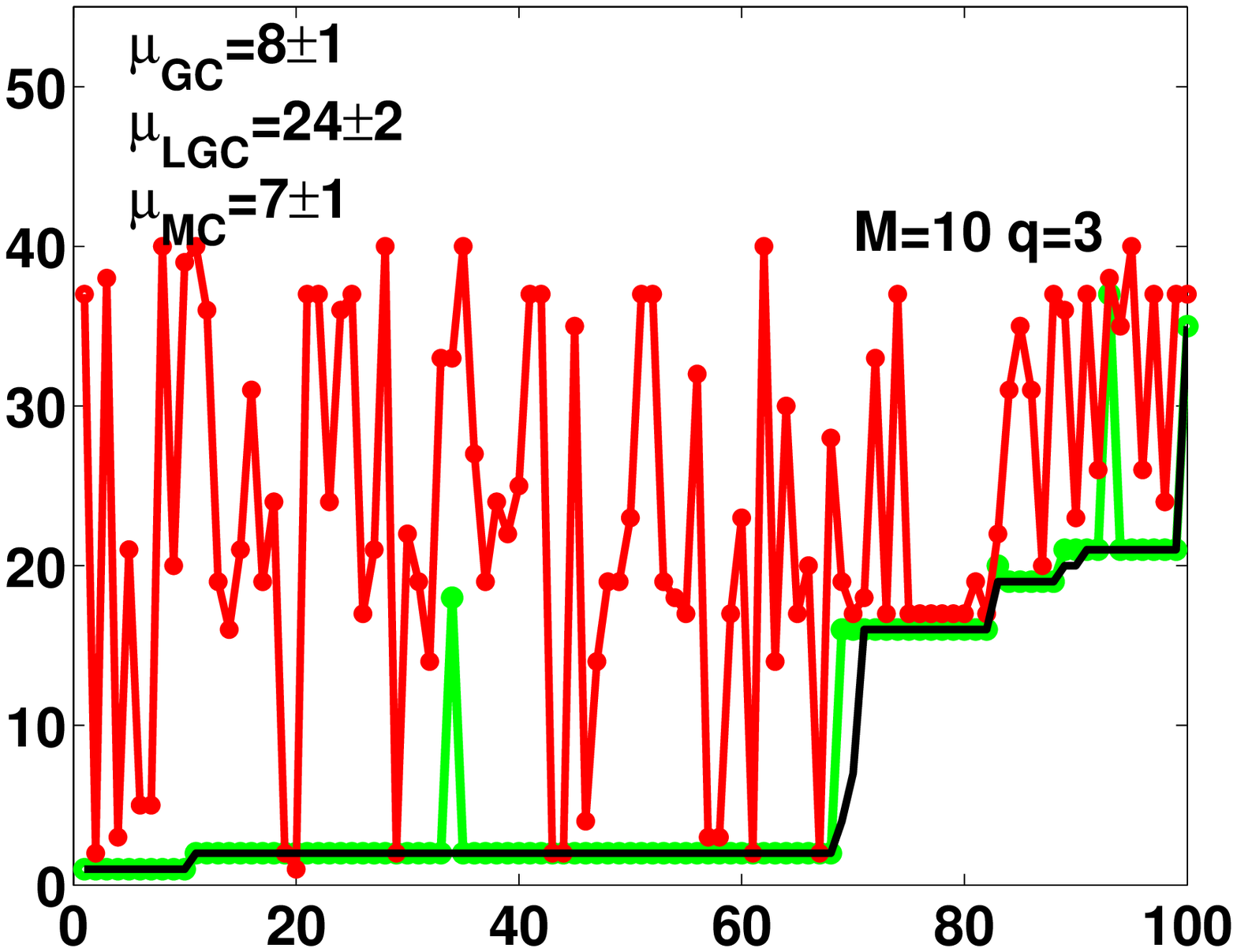,width=4cm}&
\hspace{-0.5cm}
\psfig{file=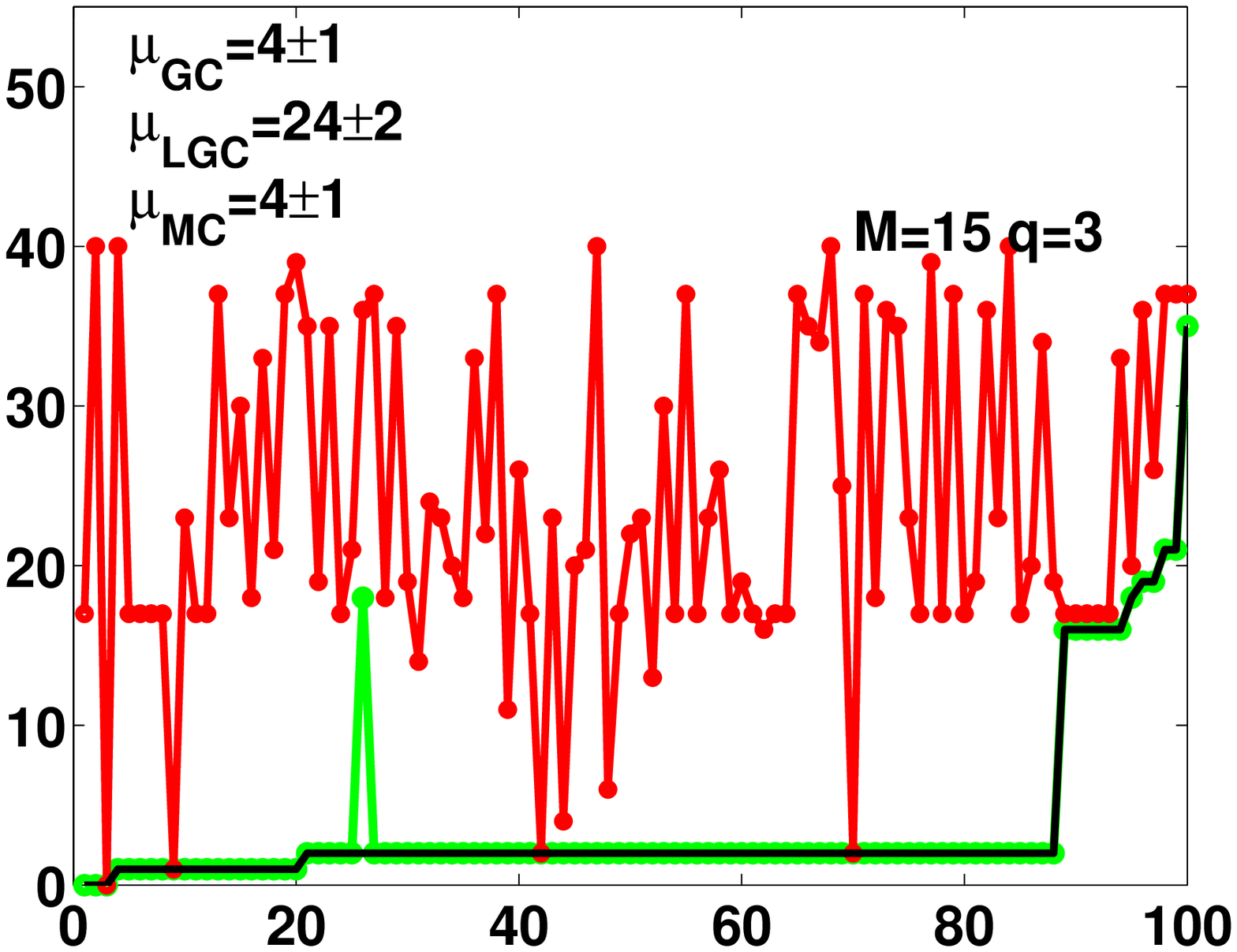,width=4cm}
\hspace{-0.5cm}
\end{tabular}
\end{center}
}\hfill
\parbox{4cm}
{
\caption{\capfont The same as in Fig.~\ref{fig:threeWavesClassification} but
for $57$ leukemia samples. The values of $q$ and $M$ appears in each panel.
The  number of misclassified points in  the unsupervised case was $11$.
}
}
\label{fig:Leukemia}
\vspace{-0.81cm}
\end{figure*}

\vspace{-0.2cm}
\subsection{Yeast cell cycle gene expression data}
\label{sec:cell_cycle}
\vspace{-0.2cm}
In this section we describe  an application of our method to a real-life
problem in cellular biology for which the true solution is partly unknown.
This concerns the assignment of the yeast's genes to the stage in the
cell cycle in which they are expressed.
While the yeast's genome is well-characterized, the function of many of its
genes remains to be determined.
Therefore, correctly  assigning genes to their cell cycle phase may shed light on
their function and help connect them to the emerging cellular network.

The Yeast's cell cycle was studied by various researchers, typically by applying
unsupervised methods, e.g.~\cite{spellman,alter}.
Here we use the data of Spellman et al. which measured the expression level
of the yeast's genes at $18$ specific times over the course of two cell-cycles,
thus data consists of $18$ measurements of more than $6000$ genes.
Due to experimental difficulties some of the entries in this $18\times 6000$
matrix are missing, hence following~\cite{alter} we used a subset of  $4523$ genes
for which at least $15$ out of the $18$ readings are available.
For $77$ of these genes, the assignment to one of  $5$ stages
in the cell cycle (M/G1, G1, S, S/G2 and G2/M) is well established.
Therefore, we have a multi-class classification problem ($q=5$)
of $4523$ points in $18$ dimensions, with $77$ labelled points.
As a similarity measure we used a  standard protocol as
 in Sec.~\ref{sec:leukemia}.

Since ground truth is not available in this problem we decided
to measure the success rate of our method by comparing our results to
the proposed classification of Spellman et al.
They used several biological criteria in order  to rank the
genes according to their participation in the cell-cycle.
Their list consists of $604$  out of the $4523$ genes, and $69$ of them
also appear in the list of known  $77$ genes, leaving $535$ genes as a test set.

  We classified the $535$ points to one of the $5$
classes, or marked them as `confused' between classes.
When considering only the classified points and
treating the `confused' points as errors
our average success rate is 32\% (over the 5 classes),
while  graph-cuts reaches 20\%.


\vspace{-0.25cm}
\section{Discussion}
\label{sec:discussion}
We introduced an approach to semi-supervised learning which is based on
statistical physics. Our approach may be applied to any
energy function, and  yields an equal or better performance
than minimizing the same energy function.
Our method is most suitable in case the number labelled points is small,
since  its classifications  would coincide with the minimal energy solution
as the number   of labelled points becomes larger.

The method is based on the Multicanonical MCMC method,
which allows for an efficient estimation of the Boltzmann distribution,
even in the multi-class scenario.
The basic difficulty in methods which seek the minimal energy, i.e. work at $T=0$,
is that the multi-class scenario is NP-hard.
We avoid such difficulties since the
 interesting regime for classification is    $T>0$.

The computational complexity of MCMC  is hard
to estimate, as it is problem dependent.
A large multi-class data set may indeed be difficult to sample,
and require a long run, which calls  for even more
efficient MCMC or approximation methods.
However, we hope to have convinced the reader
that our performance gain  over other, more efficient,
methods may be worthwhile.

Although our results display the advantages of incorporating labelled points
in an unsupervised setting, the performance highly depends on the specific
choice of labelled points, and in some cases it is even preferable
to ignore the labelled points. A related  phenomenon already appeared in previous work,
e.g.~\cite{iraCohen}, and should be thoroughly addressed.

\vspace{-0.25cm}
\section*{Acknowledgments}
\vspace{-0.25cm}
This work was partially supported by a Program Project Grant from
the National Cancer Institute (P01-CA65930).
\vspace{-0.25cm}
\bibliography{semiICML}
\vspace{-0.25cm}
\bibliographystyle{plain}

\end{document}